\documentclass[twoside]{article}

\usepackage[preprint]{aistats2027}

\usepackage[round]{natbib}

\usepackage{amsfonts}
\usepackage{amssymb}
\usepackage{amsmath}
\usepackage{amsthm}
\usepackage{algorithm}
\usepackage{algpseudocode}
\usepackage{booktabs}
\usepackage{capt-of}%
\usepackage{longtable}
\usepackage{multirow}
\usepackage[table]{xcolor}
\usepackage{graphicx}
\graphicspath{{img/}}
\usepackage{hyperref}
\usepackage{xurl}
\usepackage{pifont}
\usepackage{enumitem}

\definecolor{savpos}{RGB}{222,235,247}
\definecolor{savneg}{RGB}{253,226,200}
\definecolor{savnull}{RGB}{242,242,242}
\definecolor{gmshade}{RGB}{230,243,232}%
\newcommand{\savbenefit}[1]{\cellcolor{savpos}#1}
\newcommand{\savharm}[1]{\cellcolor{savneg}#1}
\newcommand{\savinconclusive}[1]{\cellcolor{savnull}#1}
\newtheorem{proposition}{Proposition}
\usepackage{placeins}
\newcommand{\GeoReplay}{\mbox{\textsc{GeoReplay}}}%
\providecommand{\Beta}{\beta}%

\begin{document}
\runningauthor{Nerrise, Busch, Hora, Sipka, Maddix, Brikis}

\twocolumn[

\aistatstitle{Measuring and Improving Transfer of Neural Surrogates for Vehicle Aerodynamics}

\aistatsauthor{Favour Nerrise$^{1,2}$\textsuperscript{*} \qquad Julian Busch$^{1}$ \qquad Gurpreet Singh Hora$^{1}$ \\[4pt] \bfseries Martin Sipka$^{1}$ \qquad Danielle C. Maddix$^{1}$ \qquad Georgia Olympia Brikis$^{1}$}

\aistatsaddress{\rule{0pt}{12pt}$^{1}$Siemens \qquad $^{2}$Department of Electrical Engineering, Stanford University, USA \\[3pt] \texttt{fnerrise@stanford.edu, danielle.maddix-robinson@siemens.com, georgia.brikis@siemens.com}} ]

\begingroup
\renewcommand{\thefootnote}{*}
\footnotetext{This work was conducted during an internship at Siemens.}
\endgroup

\begin{abstract}
Neural surrogates for external vehicle aerodynamics are fast alternatives to physics-based computational fluid dynamics (CFD) simulations. Their foundational limitation is a lack of zero-shot generalization across vehicle classes, which results in an expensive, custom training campaign for each new vehicle design. Similar to pretrained models in other fields, a pretrained surrogate should adapt to a new vehicle class using only a few simulations and stay accurate on its pretraining domain. No prior study has systematically tested this transfer behavior. We conduct a large-scale, controlled benchmarking study on five DrivAerML-pretrained surrogates across four independent target datasets. These datasets differ from the source in geometry, CFD solver, turbulence model, and flow conditions. We find that fine-tuning these surrogates outperforms training from scratch in fewer than half of the comparisons and lowers the accuracy of the surrogate on its source dataset. We introduce \GeoReplay{}, a geometry- and gradient-aligned source replay that keeps source simulations resembling the target in training and applies their updates only when they agree with the target gradient. A risk bound separates geometric coverage from differences in the simulated physics, and we prove that every sufficiently small \GeoReplay{} step decreases the target training objective. \GeoReplay{} lowers the source error of fine-tuning in 77 of 80 settings and matches its target accuracy. %
\end{abstract}

\begin{figure*}[t]
\centering
\includegraphics[width=\linewidth]{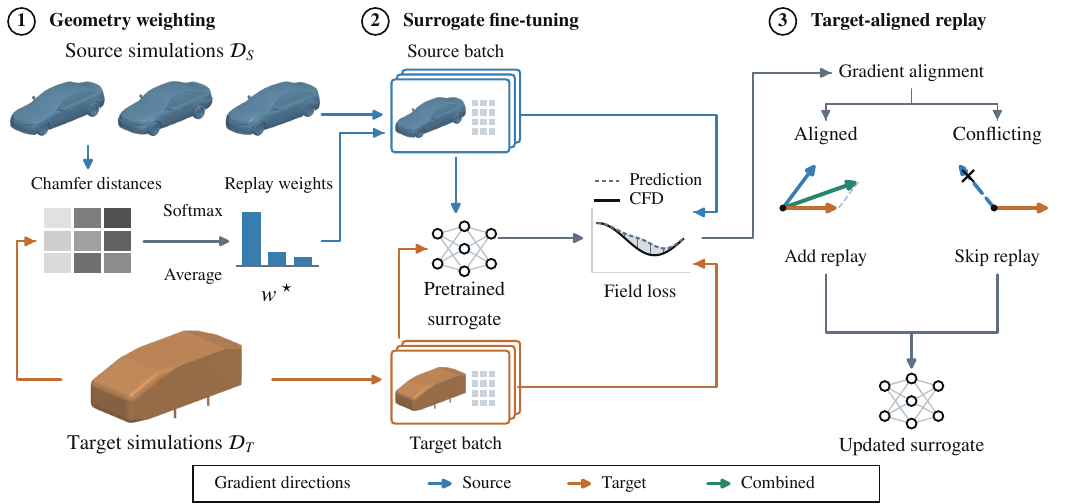}
\caption{\GeoReplay{} (1)~weights source simulations by Chamfer distance to target training geometries, (2)~fine-tunes on target and source batches, and (3)~adds a source gradient only if it aligns with the target gradient.}
\label{fig:aligned-replay}
\end{figure*}

\section{Introduction}
\label{sec:introduction}

Aerodynamic drag and lift coefficients govern vehicle range, stability, and fuel economy. Design engineers evaluate hundreds of geometries against these coefficients before converging on a design. Drag and lift are surface integrals of the pressure and wall shear stress. Computing these quantities requires numerically solving the Navier-Stokes equations around the vehicle, which is a task traditionally done by high-fidelity CFD solvers. A single simulation can take hours to converge. Neural surrogates have emerged as computationally efficient alternatives. Given a vehicle geometry, these surrogates can predict surface and volume flow fields in milliseconds to seconds \citep{pmlr-v235-wu24r,adams2025geotransolver,ranade2025domino,choy2025factorized,nabian2024x}, which makes wide design sweeps tractable \citep{ananthan2024machine,ashton2024drivaerml,ashton2024windsorml,elrefaie2024drivaernet++}. This speedup comes at an upfront cost. Each surrogate has to be trained on many CFD runs of the vehicle class, and every new class, e.g., a different body style, segment, or program, requires repeating this expensive training process. %

A pretrained aerodynamic surrogate should be able to learn the underlying physics and adapt to a new vehicle class using only a few simulations while staying accurate on its source dataset, similar to how models pretrained at scale in other domains generalize to unseen data and adapt to new tasks with little data \citep{bommasani2021opportunities}. This generalization is critical in practice since each new target domain typically comes with only a handful of CFD runs. Generating more simulations is precisely the cost that pretraining is meant to avoid \citep{breakevencomplexity}. 

Most automotive surrogates are trained and evaluated with the same vehicle dataset \citep{ashton2024drivaerml,elrefaie2024drivaernet++,tangsali2025benchmarking,elrefaie2025carbench}, whereas practitioners already apply pretrained surrogates across a broad range of geometries and application areas \citep{nvidia}. %
Although there are only a handful of publicly available CFD datasets, together they provide a diverse testbed for studying transfer across vehicle classes. These datasets span variations that a deployed surrogate would see in practice, e.g., parametrically varied notchback road cars (DrivAerML) \citep{ashton2024drivaerml}; fastback, notchback, and estateback production-style vehicles  (DrivAerNet++) \citep{elrefaie2024drivaernet++}; quarter-scale sport utility vehicles (SHIFT-SUV) \citep{shift_suv_2025}; simplified Ahmed reference bodies (AhmedML) \citep{ashton2024ahmedml}; and square-back reference bodies (WindsorML) \citep{ashton2024windsorml}.  Table~\ref{tab:datasets} summarizes how these datasets additionally differ in CFD solver, turbulence model, scale, and flow conditions.

In this work, we adopt them as our transfer learning testbed and aim to answer the following question: %
\emph{How can a pretrained aerodynamic surrogate adapt to a new vehicle
class from a few simulations while keeping its accuracy
on the source dataset?}
Answering this question raises three challenges. (i) \emph{Attribution.} Improvement after fine-tuning mixes the value of pretraining with the value of the target simulations themselves. A credible comparison must hold the architecture, data, and selection protocol fixed. (ii) \emph{Coverage versus physics.} A target geometry close to the source can still differ in solver, mesh, and flow conditions, so geometric similarity alone does not determine transfer. (iii) \emph{Conflicting source updates.} Source information that helps one target can hurt another, so adaptation must detect when a source update opposes the target objective.

\noindent\textbf{Our approach.} For (i), we compare each fine-tuned model with the same architecture trained from scratch on the same target simulations, and we evaluate both on the source and on the target (\S\ref{sec:evaluation}). For (ii), we bound the target risk by a geometric coverage term plus a physics-shift term, which explains why geometric similarity alone cannot guarantee transfer. For (iii), we introduce \GeoReplay{}, a geometry- and gradient-aligned source replay (Figure~\ref{fig:aligned-replay}, Algorithm~\ref{alg:aligned-replay}, \S\ref{sec:adaptation}). An entropy-regularized coupling weights source simulations by their proximity to the target training geometries, and a cosine gate keeps a source gradient only when it agrees with the target gradient. Across surrogates, target datasets, and budgets, \GeoReplay{} keeps the target accuracy of fine-tuning while substantially reducing its error on the source dataset (\S\ref{sec:experiments}).

In summary, our contributions are as follows.
\begin{itemize}[leftmargin=1.2em,label={\ding{68}},itemsep=2pt]
    \item We introduce \GeoReplay{}, a geometry- and gradient-aligned source replay method that adapts a pretrained surrogate to a new target while preserving its source accuracy (\S\ref{sec:adaptation}).
    \item We bound the target risk of geometry-weighted replay by separating geometric coverage from physics shift, and prove that \GeoReplay{} steps descend the target training objective (\S\ref{sec:adaptation}).
    \item We present the first large-scale study, to our knowledge, of transfer across independent automotive CFD datasets that span distinct geometries, CFD solvers, turbulence models, and Reynolds numbers, evaluating jointly on their source and target domains (\S\ref{sec:experiments}).
\end{itemize}

\section{Related Work}
\label{sec:related-work}

\begin{table*}[!t]
\centering
\caption{The source dataset and the four target datasets. Each render shows one simulation from the dataset, coloured by its time-averaged surface pressure on a scale set separately for each dataset. Beyond geometry, the datasets differ in solver, turbulence model, scale, freestream speed, and Reynolds number. Each Reynolds number uses the length scale in parentheses. -- marks unreleased values, and $^*$ are derived from the released flow conditions. Values are from the dataset papers and cards cited in Section~\ref{sec:setup}.}
\label{tab:datasets}
\footnotesize
\setlength{\tabcolsep}{4pt}
\renewcommand{\arraystretch}{1.12}
\begin{tabular}{@{}l*{5}{>{\centering\arraybackslash}p{0.148\textwidth}}@{}}
\toprule
& \textbf{DrivAerML} & \textbf{DrivAerNet++} & \textbf{SHIFT-SUV} & \textbf{AhmedML} & \textbf{WindsorML} \\
& {\scriptsize Source} & {\scriptsize Target, road car} & {\scriptsize Target, road car} & {\scriptsize Target, bluff body} & {\scriptsize Target, bluff body} \\
& \includegraphics[width=\linewidth,height=1.45cm,keepaspectratio]{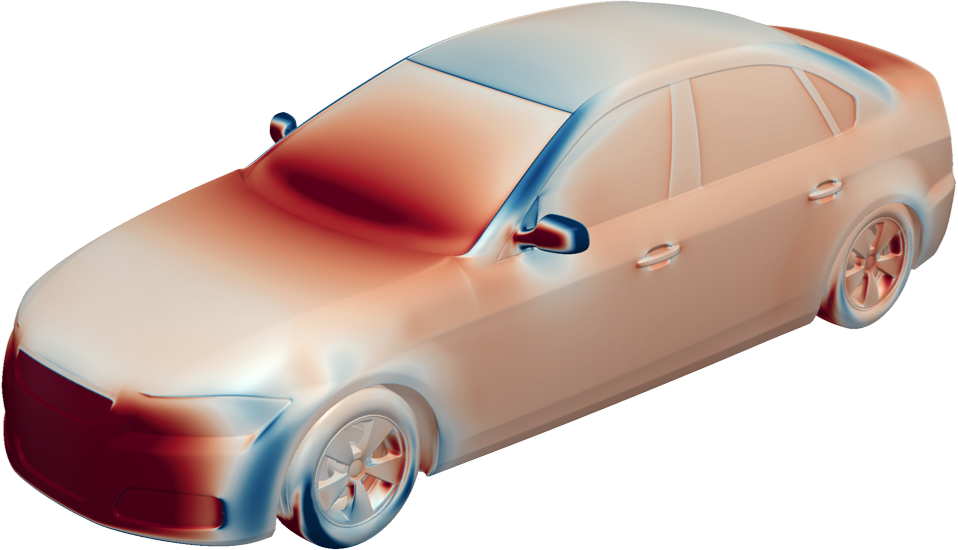} & \includegraphics[width=\linewidth,height=1.45cm,keepaspectratio]{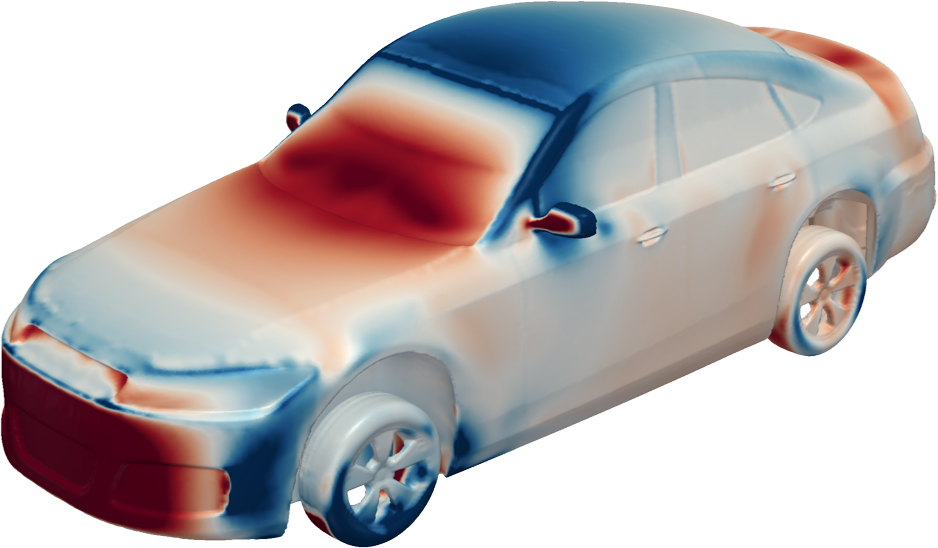} & \includegraphics[width=\linewidth,height=1.45cm,keepaspectratio]{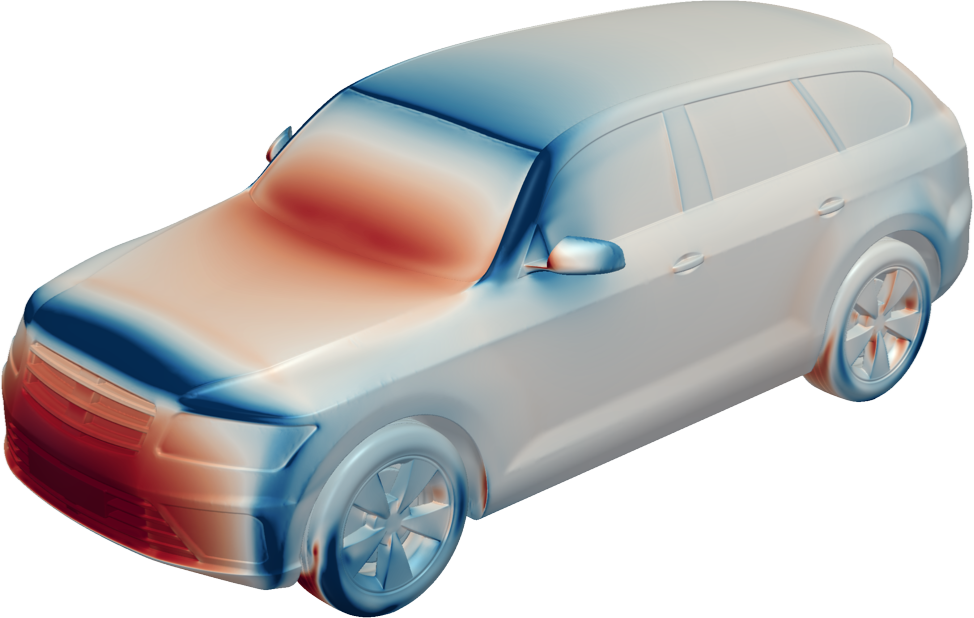} & \includegraphics[width=\linewidth,height=1.45cm,keepaspectratio]{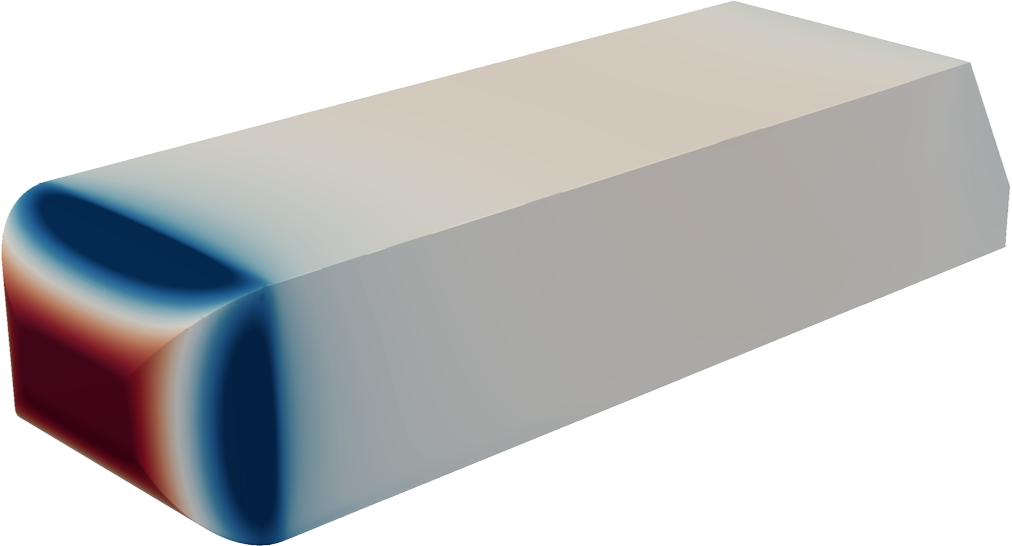} & \includegraphics[width=\linewidth,height=1.45cm,keepaspectratio]{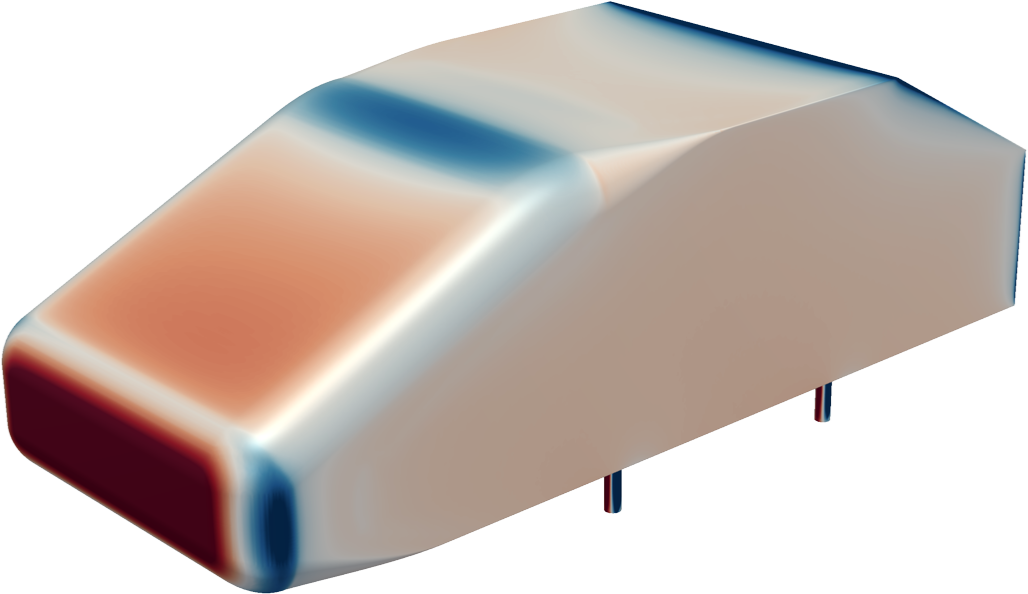} \\
\midrule
Geometry & DrivAer notchback & Fast-, notch-, estateback & AeroSUV & Ahmed body & Windsor square-back \\
Shape parameters & 16 & 26 & 10 & 6 & 7 \\
Solver & OpenFOAM v2212 & OpenFOAM v11 & Luminary Cloud & OpenFOAM v2212 & Volcano ScaLES \\
Turbulence model & SA-$\sigma$-DDES & Steady RANS, $k$-$\omega$ SST & DDES & SA-DDES & WMLES \\
Surface fields & Time-averaged & Steady & Time-averaged & Time-averaged & Time-averaged \\
Mesh cells & $\approx$160M & $\approx$24M & $\approx$45M & 15--20M & $>$280M \\
Scale & Full & Full & Quarter & Laboratory & Laboratory \\
Freestream speed & 38.9\,m/s & 30\,m/s & 50\,m/s & 1\,m/s & 40\,m/s \\
Reynolds number & $7.2\times10^6$ (wheelbase) & $8.4$--$10.1\times10^6$ (length) & $3.9\times10^6$ (length)$^*$ & $2.78\times10^6$ (length) & $2.9\times10^6$ (length) \\
Released simulations & 500 & 8,000 & 99 (sample) & 500 & 350 \\
\bottomrule
\end{tabular}
\end{table*}

\noindent\textbf{Aerodynamic surrogate benchmarks.} Prior studies do not measure transfer between independent CFD campaigns. Public CFD benchmarks \citep{ashton2024drivaerml,ashton2024ahmedml,ashton2024windsorml,elrefaie2024drivaernet++,shift_suv_2025} mostly assess surrogates within one dataset \citep{tangsali2025benchmarking} or under category shifts inside it. For example, CarBench \citep{elrefaie2025carbench} benchmarks various surrogates on different car geometries in DrivAerNet++. A limitation of this study is that it does not consider changes in the numerical solver or turbulence modeling. Surrogate architectures also continue to evolve, for example toward predicting from raw point-cloud geometry without a simulation mesh \citep{hagnberger2026smart}.

\noindent\textbf{Negative transfer and transferability.} Negative transfer is typically assessed by comparing an adapted model with a target-only baseline \citep{wang2019characterizing,hanneke2019value}, and metrics fitted from target labels can estimate transferability before adaptation \citep{nguyen2020leep,regression_transferability}. Continual learning limits forgetting of the earlier task by regularizing toward the pretrained parameters, as in EWC \citep{kirkpatrick2017overcoming} and L2-SP \citep{li2018explicit}, or by replaying earlier data, as in experience replay \citep{rolnick2019experience} and A-GEM \citep{chaudhry2019agem}.

\noindent\textbf{Transfer of scientific surrogates.} For scientific surrogates, transfer depends on how closely source and target physics match, and neural operators \citep{li2021fourier} degrade out of distribution \citep{subramanian2023towards,multiple_physics,mouli2024using}; parameter-efficient methods adapt PDE surrogates on lower-dimensional benchmarks instead of three-dimensional vehicle aerodynamics \citep{pmlr-v235-berman24b,cao2025spectral}. Cross-domain pretraining of steady-state CFD surrogates over geometries, boundary conditions and simulation fidelities lowers the error of fine-tuned models relative to training from scratch \citep{zhou2026crossdomain}; Appendix~\ref{app:pretraining} compares surrogates trained on a mixture of the five vehicle datasets with surrogates trained on each dataset alone.

\section{Problem Definition}
\label{sec:evaluation}

Let $\mathcal D_S$ and $\mathcal D_T$ denote the source and target datasets, respectively. For an architecture $f_\theta$, we compare a model initialized from parameters $\theta_{\mathrm{src}}$ learned on $\mathcal D_S$ with the same architecture initialized randomly. Both models train on $\Beta$ simulations from $\mathcal D_T$ and use the same validation simulations for checkpoint selection. We evaluate both models on the same held-out simulations. We also evaluate the source-pretrained model directly on the target test simulations (zero-shot prediction). We use zero-shot error to measure initial generalization, its reduction after fine-tuning to measure adaptation, and the fine-tuned--from-scratch comparison to measure the realized value of pretraining at fixed $\Beta$. For a method M, a subscript denotes the training dataset and a superscript the test dataset. The errors $e_S$ and $e_T$ denote the errors of $\mathrm{M}_T^S$ and $\mathrm{M}_T^T$, respectively. Since a reusable surrogate must stay accurate on both datasets, we propose to use the \emph{geometric mean} $\sqrt{e_S\,e_T}$ of both errors, which is low only when both errors are.

\paragraph{Distance to the closest training geometry.} We follow the evaluation structure
from recent automotive aerodynamic analyses that use Chamfer distance to quantify geometric diversity and relate prediction error to the closest training geometry \citep{physicsxlpm}. Let $P(g)$ contain area-weighted surface points from geometry $g$ after reconciling dataset axes, centering the body, and uniformly normalizing all coordinates by body length. For two geometries, we compute the symmetric Chamfer distance
\begin{equation}
\begin{aligned}
d_{\mathrm C}(g,g')=\frac{1}{2}\bigg[&\frac{1}{|P(g)|}
\sum_{x\in P(g)}\min_{y\in P(g')}\|x-y\|_2 \\
&+\frac{1}{|P(g')|}\sum_{y\in P(g')}
\min_{x\in P(g)}\|y-x\|_2\bigg].
\end{aligned}
\label{eq:chamfer}
\end{equation}
For source DrivAerML training geometries $\mathcal S$, the distance to the closest training geometry is $d_{\mathcal S}(g)=\min_{g'\in\mathcal S}d_{\mathrm C}(g,g')$. It is dimensionless and requires no target CFD field. We use all unordered pairs within each dataset to characterize internal geometric diversity. We separately compare all DrivAerML--target pairs. For each fixed target test geometry, we compare its distance to the closest DrivAerML pretraining geometry with its distance to the closest geometry in the corresponding target training split. We measure this distance before any training. See Appendix~\ref{app:geometry} for the computation details.

\section{\GeoReplay{}: Geometry- and Gradient-Aligned Source Replay}
\label{sec:adaptation}

We observe two obstacles to reliable transfer. The distance to the source geometry set can characterize a model's initial geometric extrapolation, but it cannot determine whether pretraining remains useful after fine-tuning. Replaying arbitrary source simulations during adaptation can bias training toward source-specific solvers, meshes, operating conditions, or turbulence models. We use geometry to decide which source simulations are relevant and the target objective to decide whether their gradients are useful.

\subsection{Geometry-Weighted Source Distribution}

We denote a simulation as $z=(g,y)$, where $g$ denotes a surface geometry and $y$ contains its CFD fields. Let $\widehat P_T^G=\Beta^{-1}\sum_{i=1}^{\Beta}\delta_{g_i^T}$ be the empirical target-geometry distribution, let $C_{ij}=d_{\mathrm C}(g_i^T,g_j^S)$, and let $u_j=1/n_S$. We compute an entropy-regularized target-to-source coupling
\begin{equation}
\begin{aligned}
\Pi^\star&=\arg\min_{\substack{\Pi\geq0\\\Pi\mathbf 1=\Beta^{-1}\mathbf 1}}\langle C,\Pi\rangle+\tau D_{\mathrm{KL}}\!\left(\Pi\middle\|\Beta^{-1}\mathbf 1u^\top\right),\\
w^\star&=\Pi^{\star\top}\mathbf 1,
\end{aligned}
\label{eq:geometry-weights}
\end{equation}
where $d_{\mathrm C}$ denotes the body-length-normalized Chamfer distance from Equation~\ref{eq:chamfer} and $\tau>0$ controls concentration. This entropy-regularized transport objective \citep{cuturi2013sinkhorn} has the row-wise solution $\Pi^\star_{ij}=\Beta^{-1}\exp(-C_{ij}/\tau)/\sum_k\exp(-C_{ik}/\tau)$. Its source marginal $w^\star$ gives every target geometry equal mass while favoring nearby source geometries. Equation~\ref{eq:geometry-weights} uses the target training geometries but does not inspect their CFD fields.

\begin{figure*}[!t]
\centering
\includegraphics[width=0.85\linewidth]{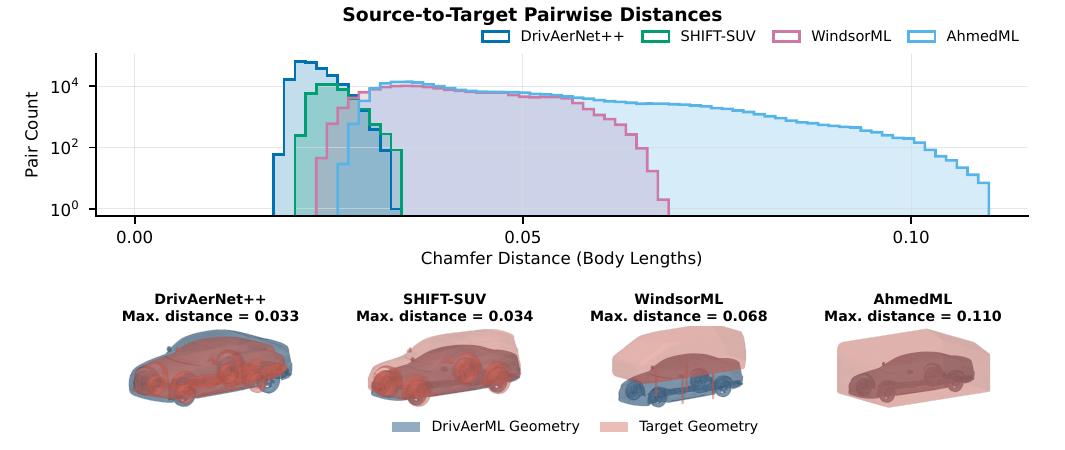}
\caption{Source-to-target pairwise geometric distances between every DrivAerML training geometry and every usable geometry in each target dataset. The most distant DrivAerML--target pair for each target dataset, ordered by increasing distance.}
\label{fig:geometry-summary}
\end{figure*}

\subsection{Target-Aligned Replay Updates}

Let $F_T(\theta)$ denote the target empirical risk and let $F_{S,w^\star}(\theta)$ denote the source risk under $w^\star$. At one optimization step, we compute a target gradient $g_T$ and a geometry-weighted source gradient $g_S$. We retain only the component of the source update that agrees with the target descent direction:
\begin{equation}
\begin{aligned}
a(g_T,g_S)&=\max\!\left(0,\frac{\langle g_T,g_S\rangle}{\lVert g_T\rVert\lVert g_S\rVert+\epsilon}\right),\\
g_R&=a(g_T,g_S)\frac{\lVert g_T\rVert}{\lVert g_S\rVert+\epsilon}g_S.
\end{aligned}
\label{eq:aligned-gradient}
\end{equation}
The update is $\theta^+=\theta-\eta(g_T+\lambda g_R)$, where $\lambda\geq0$ sets the replay strength. The cosine gate in Equation~\ref{eq:aligned-gradient} is the gradient-similarity rule of \citet{du2018adapting} for auxiliary losses, to which we add a rescaling to the target gradient norm. Unlike gradient surgery \citep{yu2020gradient}, it leaves the target gradient unchanged. GEM and A-GEM \citep{lopezpaz2017gem,chaudhry2019agem} constrain each update with the gradient of an episodic memory of earlier-task samples; \GeoReplay{} uses the source simulations as this memory, samples them by geometry, and drops conflicting source gradients instead of projecting. If no sampled source batch has positive alignment, $g_R=0$ and the method reduces to ordinary target fine-tuning. We summarize the full \GeoReplay{} procedure in Algorithm~\ref{alg:aligned-replay}.

\begin{algorithm}[t]
\caption{\GeoReplay{}}
\label{alg:aligned-replay}
\begin{algorithmic}[1]
\Require $\theta_{\mathrm{src}}$, $\mathcal D_S$, $\mathcal D_T$, replay strength $\lambda$, entropy weight $\tau$
\State Compute pairwise geometry costs $d_{\mathrm C}(g_i^T,g_j^S)$
\State Solve Equation~\ref{eq:geometry-weights} for source weights $w^\star$
\State $\theta\gets\theta_{\mathrm{src}}$
\While{the target validation criterion improves}
  \State Sample a target batch $\mathcal M_T$ and a source batch $\mathcal M_S\sim w^\star$
  \State $g_T\gets\nabla_\theta \widehat F_{\mathcal M_T}(\theta)$; $g_S\gets\nabla_\theta \widehat F_{\mathcal M_S}(\theta)$
  \State Compute $g_R$ using Equation~\ref{eq:aligned-gradient}
  \State $\theta\gets\theta-\eta(g_T+\lambda g_R)$
\EndWhile
\State \Return the checkpoint with minimum target validation error
\end{algorithmic}
\end{algorithm}

\subsection{What the Two Alignment Steps Guarantee}
The geometry weights control one identifiable part of source--target discrepancy. They cannot remove changes in the conditional CFD solution induced by a different solver, mesh, operating condition, or turbulence model. Proposition~\ref{prop:risk-bound} makes this separation explicit. For a fixed $\theta$, define the conditional risks $m_D(g)=\mathbb E_D[\ell(f_\theta(g),Y)\mid G=g]$ for $D\in\{S,T\}$.

\begin{proposition}[Geometry-conditioned risk discrepancy]
\label{prop:risk-bound}
Assume that $|m_T(g)-m_T(g')|\leq L_Gd_{\mathrm C}(g,g')$ and that $|m_T(g)-m_S(g)|\leq\delta_{\mathrm{phys}}$ on the support of $Q_w^G$. Let $\mathcal T_{d_{\mathrm C}}(P,Q)=\inf_{\pi\in\Pi(P,Q)}\mathbb E_{(G,G')\sim\pi}[d_{\mathrm C}(G,G')]$. Then
\begin{equation}
\left|R_T(\theta)-R_{S,w}(\theta)\right|\leq L_G\mathcal T_{d_{\mathrm C}}(P_T^G,Q_w^G)+\delta_{\mathrm{phys}}.
\label{eq:risk-bound}
\end{equation}
\end{proposition}

The first term decreases when the weighted source geometries cover the target geometries. The residual $\delta_{\mathrm{phys}}$ includes conditional differences that geometry alone cannot identify. This term explains why geometric proximity can coexist with negative transfer. Proposition~\ref{prop:risk-bound} is an instance of the Wasserstein-type domain adaptation bounds of \citet{redko2017theoretical}, \citet{courty2017joint}, and \citet{shen2018wasserstein}, with the Chamfer distance as ground metric and the physics shift as the residual term.

\begin{proposition}[Target-objective descent]
\label{prop:descent}
Assume that $F_T$ is $L$-smooth, $g_T=\nabla F_T(\theta)$, and $\langle g_T,g_R\rangle\geq0$. Let $d=g_T+\lambda g_R$. For any
\begin{equation}
0<\eta\leq\frac{2\langle g_T,d\rangle}{L\lVert d\rVert^2},
\label{eq:step-condition}
\end{equation}
the update $\theta^+=\theta-\eta d$ satisfies $F_T(\theta^+)\leq F_T(\theta)$.
\end{proposition}

The cosine gate enforces the alignment condition by construction. Proposition~\ref{prop:descent} is a one-step statement about the target training objective; it does not guarantee lower held-out error or improvement over ordinary fine-tuning. See Appendix~\ref{app:proofs} for the proofs.

\begin{table*}[!t]
\centering
\caption{Pressure errors of surrogates pretrained on DrivAerML ($S$) and adapted with $\Beta=10$ training simulations of each target dataset $T$. The left block gives the source error $e_S$ of $\mathrm{M}_T^S$, the middle block the target error $e_T$ of $\mathrm{M}_T^T$, and the shaded block their geometric mean $\sqrt{e_S\,e_T}$. ZS is the pretrained model without adaptation, so its $e_S$ does not depend on $T$. SC trains from scratch on $T$, FT fine-tunes, GR is \GeoReplay{}, and sGR is sparse \GeoReplay{}. In each block, \textbf{bold} marks the lowest value in each row. See App.~\ref{app:mean_var} and~\ref{app:georeplay-budget} for the standard deviation over three seeds and for other quantities and budgets.}
\label{tab:main-b10}
\footnotesize
\setlength{\tabcolsep}{3pt}
\resizebox{\textwidth}{!}{%
\begin{tabular}{llrrrrrrrrrrrrrrr}
\toprule
 & & \multicolumn{5}{c}{Source error $e_S$} & \multicolumn{5}{c}{Target error $e_T$} & \multicolumn{5}{c}{Geometric mean $\sqrt{e_S\,e_T}$} \\
\cmidrule(lr){3-7}\cmidrule(lr){8-12}\cmidrule(lr){13-17}
$T$ & Surrogate & ZS $\downarrow$ & SC $\downarrow$ & FT $\downarrow$ & GR $\downarrow$ & sGR $\downarrow$ & ZS $\downarrow$ & SC $\downarrow$ & FT $\downarrow$ & GR $\downarrow$ & sGR $\downarrow$ & ZS $\downarrow$ & SC $\downarrow$ & FT $\downarrow$ & GR $\downarrow$ & sGR $\downarrow$ \\
\midrule
\multirow{5}{*}{\rotatebox[origin=c]{90}{\shortstack{DrivAer\\Net++}}} & TSol & \textbf{0.042} & 0.736 & 0.350 & 0.096 & 0.290 & 1.020 & 0.215 & \textbf{0.197} & \textbf{0.197} & \textbf{0.197} & \cellcolor{gmshade}0.207 & \cellcolor{gmshade}0.398 & \cellcolor{gmshade}0.263 & \cellcolor{gmshade}\textbf{0.137} & \cellcolor{gmshade}0.239 \\
 & GeoT & \textbf{0.037} & 0.764 & 0.199 & 0.060 & 0.063 & 1.271 & \textbf{0.202} & 0.205 & 0.206 & 0.208 & \cellcolor{gmshade}0.217 & \cellcolor{gmshade}0.393 & \cellcolor{gmshade}0.202 & \cellcolor{gmshade}\textbf{0.111} & \cellcolor{gmshade}0.114 \\
 & DoM & 0.186 & 0.830 & 0.461 & \textbf{0.135} & 0.232 & 0.794 & 0.242 & \textbf{0.214} & 0.215 & \textbf{0.214} & \cellcolor{gmshade}0.384 & \cellcolor{gmshade}0.448 & \cellcolor{gmshade}0.314 & \cellcolor{gmshade}\textbf{0.170} & \cellcolor{gmshade}0.223 \\
 & XMGN & \textbf{0.278} & 0.719 & 0.499 & 0.392 & 0.371 & 1.194 & 0.234 & 0.226 & \textbf{0.221} & \textbf{0.221} & \cellcolor{gmshade}0.576 & \cellcolor{gmshade}0.410 & \cellcolor{gmshade}0.336 & \cellcolor{gmshade}0.294 & \cellcolor{gmshade}\textbf{0.286} \\
 & GINO & \textbf{0.268} & 0.777 & 0.555 & 0.360 & 0.432 & 0.865 & \textbf{0.283} & \textbf{0.283} & 0.286 & 0.284 & \cellcolor{gmshade}0.481 & \cellcolor{gmshade}0.469 & \cellcolor{gmshade}0.397 & \cellcolor{gmshade}\textbf{0.321} & \cellcolor{gmshade}0.350 \\
\midrule
\multirow{5}{*}{\rotatebox[origin=c]{90}{\shortstack{SHIFT-\\SUV}}} & TSol & \textbf{0.042} & 0.722 & 0.414 & 0.080 & 0.290 & 0.733 & 0.103 & \textbf{0.097} & 0.098 & 0.098 & \cellcolor{gmshade}0.175 & \cellcolor{gmshade}0.272 & \cellcolor{gmshade}0.200 & \cellcolor{gmshade}\textbf{0.089} & \cellcolor{gmshade}0.168 \\
 & GeoT & \textbf{0.037} & 0.739 & 0.236 & 0.057 & 0.057 & 0.845 & \textbf{0.067} & 0.081 & 0.082 & 0.082 & \cellcolor{gmshade}0.177 & \cellcolor{gmshade}0.223 & \cellcolor{gmshade}0.138 & \cellcolor{gmshade}\textbf{0.068} & \cellcolor{gmshade}0.069 \\
 & DoM & 0.186 & 1.261 & 0.485 & \textbf{0.134} & 0.167 & 0.687 & 0.134 & \textbf{0.119} & 0.124 & 0.120 & \cellcolor{gmshade}0.358 & \cellcolor{gmshade}0.411 & \cellcolor{gmshade}0.240 & \cellcolor{gmshade}\textbf{0.129} & \cellcolor{gmshade}0.141 \\
 & XMGN & \textbf{0.278} & 0.814 & 0.714 & 0.445 & 0.396 & 1.110 & \textbf{0.103} & 0.117 & 0.111 & 0.111 & \cellcolor{gmshade}0.556 & \cellcolor{gmshade}0.289 & \cellcolor{gmshade}0.289 & \cellcolor{gmshade}0.222 & \cellcolor{gmshade}\textbf{0.210} \\
 & GINO & \textbf{0.268} & 0.720 & 0.972 & 0.428 & 0.509 & 0.582 & \textbf{0.221} & 0.239 & 0.239 & 0.239 & \cellcolor{gmshade}0.395 & \cellcolor{gmshade}0.399 & \cellcolor{gmshade}0.482 & \cellcolor{gmshade}\textbf{0.320} & \cellcolor{gmshade}0.349 \\
\midrule
\multirow{5}{*}{\rotatebox[origin=c]{90}{AhmedML}} & TSol & \textbf{0.042} & 0.824 & 0.272 & 0.075 & 0.211 & 1.041 & \textbf{0.136} & 0.165 & 0.170 & 0.165 & \cellcolor{gmshade}0.209 & \cellcolor{gmshade}0.335 & \cellcolor{gmshade}0.212 & \cellcolor{gmshade}\textbf{0.113} & \cellcolor{gmshade}0.187 \\
 & GeoT & \textbf{0.037} & 0.809 & 0.226 & 0.058 & 0.066 & 0.988 & \textbf{0.124} & 0.148 & 0.149 & 0.145 & \cellcolor{gmshade}0.192 & \cellcolor{gmshade}0.317 & \cellcolor{gmshade}0.182 & \cellcolor{gmshade}\textbf{0.093} & \cellcolor{gmshade}0.098 \\
 & DoM & 0.186 & 1.306 & 0.286 & \textbf{0.146} & 0.184 & 1.242 & \textbf{0.243} & 0.256 & 0.253 & 0.254 & \cellcolor{gmshade}0.481 & \cellcolor{gmshade}0.563 & \cellcolor{gmshade}0.271 & \cellcolor{gmshade}\textbf{0.192} & \cellcolor{gmshade}0.216 \\
 & XMGN & \textbf{0.278} & 0.941 & 0.396 & 0.332 & 0.352 & 2.036 & 0.329 & 0.292 & \textbf{0.288} & 0.297 & \cellcolor{gmshade}0.753 & \cellcolor{gmshade}0.557 & \cellcolor{gmshade}0.340 & \cellcolor{gmshade}\textbf{0.309} & \cellcolor{gmshade}0.323 \\
 & GINO & \textbf{0.268} & 1.707 & 0.579 & 0.270 & 0.277 & 1.043 & 0.590 & 0.486 & 0.474 & \textbf{0.470} & \cellcolor{gmshade}0.528 & \cellcolor{gmshade}1.003 & \cellcolor{gmshade}0.530 & \cellcolor{gmshade}\textbf{0.358} & \cellcolor{gmshade}0.361 \\
\midrule
\multirow{5}{*}{\rotatebox[origin=c]{90}{WindsorML}} & TSol & \textbf{0.042} & 1.227 & 0.333 & 0.076 & 0.310 & 0.794 & 0.427 & 0.255 & \textbf{0.253} & 0.256 & \cellcolor{gmshade}0.182 & \cellcolor{gmshade}0.724 & \cellcolor{gmshade}0.292 & \cellcolor{gmshade}\textbf{0.139} & \cellcolor{gmshade}0.282 \\
 & GeoT & \textbf{0.037} & 0.805 & 0.223 & 0.058 & 0.061 & 0.744 & \textbf{0.173} & 0.244 & 0.237 & 0.237 & \cellcolor{gmshade}0.166 & \cellcolor{gmshade}0.373 & \cellcolor{gmshade}0.233 & \cellcolor{gmshade}\textbf{0.117} & \cellcolor{gmshade}0.120 \\
 & DoM & 0.186 & 1.808 & 0.296 & \textbf{0.144} & 0.169 & 1.519 & \textbf{0.279} & 0.295 & 0.291 & 0.294 & \cellcolor{gmshade}0.532 & \cellcolor{gmshade}0.710 & \cellcolor{gmshade}0.295 & \cellcolor{gmshade}\textbf{0.205} & \cellcolor{gmshade}0.223 \\
 & XMGN & \textbf{0.278} & 0.870 & 0.451 & 0.348 & 0.377 & 0.900 & 0.329 & \textbf{0.322} & 0.324 & 0.323 & \cellcolor{gmshade}0.500 & \cellcolor{gmshade}0.535 & \cellcolor{gmshade}0.381 & \cellcolor{gmshade}\textbf{0.335} & \cellcolor{gmshade}0.349 \\
 & GINO & \textbf{0.268} & 0.901 & 0.867 & 0.271 & 0.280 & 0.790 & \textbf{0.470} & 0.487 & 0.484 & 0.484 & \cellcolor{gmshade}0.460 & \cellcolor{gmshade}0.651 & \cellcolor{gmshade}0.650 & \cellcolor{gmshade}\textbf{0.362} & \cellcolor{gmshade}0.368 \\
\bottomrule
\end{tabular}}

\end{table*}

\begin{table*}[!t]
\centering
\caption{Pressure errors for \GeoReplay{} component ablation at $\Beta=10$. The replay variants cross uniform or geometry-weighted source sampling with a plain update $g_T+\lambda g_S$ or the aligned update of Equation~\ref{eq:aligned-gradient}. A-GEM \citep{chaudhry2019agem}, L2-SP \citep{li2018explicit}, and EWC \citep{kirkpatrick2017overcoming} are continual-learning baselines; L2-SP and EWC use the regularization strength in $\{0.01, 0.1, 1\}$ with the lowest target validation error. Errors $e_S$, $e_T$, and their geometric mean (shaded) are as in Table~\ref{tab:main-b10}. Within each surrogate, \textbf{bold} marks the lowest value in each column and \underline{underline} the second lowest.}
\label{tab:ablation-b10}
\footnotesize
\setlength{\tabcolsep}{3pt}
\resizebox{\textwidth}{!}{%
\begin{tabular}{lllrrrrrrrrrrrr}
\toprule
 & & & \multicolumn{3}{c}{DrivAerNet++} & \multicolumn{3}{c}{SHIFT-SUV} & \multicolumn{3}{c}{AhmedML} & \multicolumn{3}{c}{WindsorML} \\
\cmidrule(lr){4-6}\cmidrule(lr){7-9}\cmidrule(lr){10-12}\cmidrule(lr){13-15}
Variant & Weights & Update & $e_S\downarrow$ & $e_T\downarrow$ & $\sqrt{e_S e_T}\downarrow$ & $e_S\downarrow$ & $e_T\downarrow$ & $\sqrt{e_S e_T}\downarrow$ & $e_S\downarrow$ & $e_T\downarrow$ & $\sqrt{e_S e_T}\downarrow$ & $e_S\downarrow$ & $e_T\downarrow$ & $\sqrt{e_S e_T}\downarrow$ \\
\midrule
\multicolumn{15}{l}{\textit{Transolver}} \\
FT (no replay) & -- & -- & 0.350 & \textbf{0.197} & \cellcolor{gmshade}0.263 & 0.414 & \textbf{0.097} & \cellcolor{gmshade}0.200 & 0.272 & 0.165 & \cellcolor{gmshade}0.212 & 0.333 & 0.255 & \cellcolor{gmshade}0.292 \\
Plain replay & Uniform & Plain & \underline{0.072} & 0.200 & \cellcolor{gmshade}\textbf{0.120} & \underline{0.070} & 0.113 & \cellcolor{gmshade}0.089 & \textbf{0.071} & 0.182 & \cellcolor{gmshade}0.114 & \textbf{0.071} & \underline{0.254} & \cellcolor{gmshade}\textbf{0.134} \\
Weighting only & Geometry & Plain & \textbf{0.071} & 0.201 & \cellcolor{gmshade}\textbf{0.120} & \textbf{0.069} & 0.113 & \cellcolor{gmshade}\underline{0.088} & \underline{0.074} & 0.186 & \cellcolor{gmshade}0.117 & \underline{0.074} & 0.263 & \cellcolor{gmshade}\underline{0.139} \\
Alignment only & Uniform & Aligned & 0.098 & \textbf{0.197} & \cellcolor{gmshade}0.139 & 0.079 & \textbf{0.097} & \cellcolor{gmshade}\textbf{0.087} & 0.076 & 0.162 & \cellcolor{gmshade}\textbf{0.111} & 0.076 & 0.261 & \cellcolor{gmshade}0.141 \\
A-GEM & Uniform & Projection & 0.132 & \underline{0.198} & \cellcolor{gmshade}0.161 & 0.086 & \textbf{0.097} & \cellcolor{gmshade}0.091 & 0.077 & \textbf{0.159} & \cellcolor{gmshade}\textbf{0.111} & 0.079 & 0.258 & \cellcolor{gmshade}0.143 \\
L2-SP & -- & Anchor & 0.457 & 0.219 & \cellcolor{gmshade}0.316 & 0.424 & 0.157 & \cellcolor{gmshade}0.258 & 0.242 & 0.181 & \cellcolor{gmshade}0.209 & 0.381 & 0.265 & \cellcolor{gmshade}0.318 \\
EWC & -- & Fisher anchor & 0.275 & 0.200 & \cellcolor{gmshade}0.235 & 0.204 & 0.112 & \cellcolor{gmshade}0.151 & 0.105 & \underline{0.160} & \cellcolor{gmshade}0.130 & 0.255 & 0.258 & \cellcolor{gmshade}0.257 \\
\GeoReplay{} & Geometry & Aligned & 0.096 & \textbf{0.197} & \cellcolor{gmshade}\underline{0.137} & 0.080 & \underline{0.098} & \cellcolor{gmshade}0.089 & 0.075 & 0.170 & \cellcolor{gmshade}\underline{0.113} & 0.076 & \textbf{0.253} & \cellcolor{gmshade}\underline{0.139} \\
\midrule
\multicolumn{15}{l}{\textit{GeoTransolver}} \\
FT (no replay) & -- & -- & 0.199 & \textbf{0.205} & \cellcolor{gmshade}0.202 & 0.236 & \underline{0.081} & \cellcolor{gmshade}0.138 & 0.226 & \underline{0.148} & \cellcolor{gmshade}0.182 & 0.223 & 0.244 & \cellcolor{gmshade}0.233 \\
Plain replay & Uniform & Plain & \textbf{0.056} & 0.208 & \cellcolor{gmshade}\underline{0.108} & \textbf{0.054} & 0.087 & \cellcolor{gmshade}\underline{0.069} & \textbf{0.055} & 0.152 & \cellcolor{gmshade}\underline{0.092} & \textbf{0.055} & 0.240 & \cellcolor{gmshade}\textbf{0.115} \\
Weighting only & Geometry & Plain & \textbf{0.056} & \textbf{0.205} & \cellcolor{gmshade}\textbf{0.107} & \textbf{0.054} & 0.086 & \cellcolor{gmshade}\underline{0.069} & \textbf{0.055} & \textbf{0.145} & \cellcolor{gmshade}\textbf{0.090} & \underline{0.056} & \underline{0.237} & \cellcolor{gmshade}\textbf{0.115} \\
Alignment only & Uniform & Aligned & \underline{0.059} & \textbf{0.205} & \cellcolor{gmshade}0.110 & \underline{0.056} & \underline{0.081} & \cellcolor{gmshade}\textbf{0.068} & \underline{0.056} & 0.155 & \cellcolor{gmshade}0.093 & 0.058 & \textbf{0.236} & \cellcolor{gmshade}\underline{0.117} \\
A-GEM & Uniform & Projection & 0.063 & 0.207 & \cellcolor{gmshade}0.114 & 0.058 & \textbf{0.080} & \cellcolor{gmshade}\textbf{0.068} & 0.060 & 0.156 & \cellcolor{gmshade}0.097 & 0.060 & 0.242 & \cellcolor{gmshade}0.121 \\
L2-SP & -- & Anchor & 0.253 & 0.227 & \cellcolor{gmshade}0.240 & 0.353 & 0.143 & \cellcolor{gmshade}0.225 & 0.188 & 0.173 & \cellcolor{gmshade}0.180 & 0.258 & 0.248 & \cellcolor{gmshade}0.253 \\
EWC & -- & Fisher anchor & 0.098 & 0.209 & \cellcolor{gmshade}0.143 & 0.091 & 0.094 & \cellcolor{gmshade}0.093 & 0.059 & 0.163 & \cellcolor{gmshade}0.098 & 0.089 & 0.241 & \cellcolor{gmshade}0.146 \\
\GeoReplay{} & Geometry & Aligned & 0.060 & \underline{0.206} & \cellcolor{gmshade}0.111 & 0.057 & 0.082 & \cellcolor{gmshade}\textbf{0.068} & 0.058 & 0.149 & \cellcolor{gmshade}0.093 & 0.058 & \underline{0.237} & \cellcolor{gmshade}\underline{0.117} \\
\bottomrule
\end{tabular}}

\end{table*}

\paragraph{Complexity.}
Constructing the geometry cost matrix requires $O(\Beta n_SC_d)$ work, where $C_d$ denotes the cost of one geometry comparison, and normalizing the coupling requires $O(\Beta n_S)$ work. These computations occur once before adaptation. Each training step adds one source forward and backward pass. The gradient gate requires $O(p)$ operations for $p$ trainable parameters. Source and target gradients can be accumulated sequentially, so the method does not need to store both computation graphs at once.

\begin{figure*}[t]
\centering
\includegraphics[width=\linewidth]{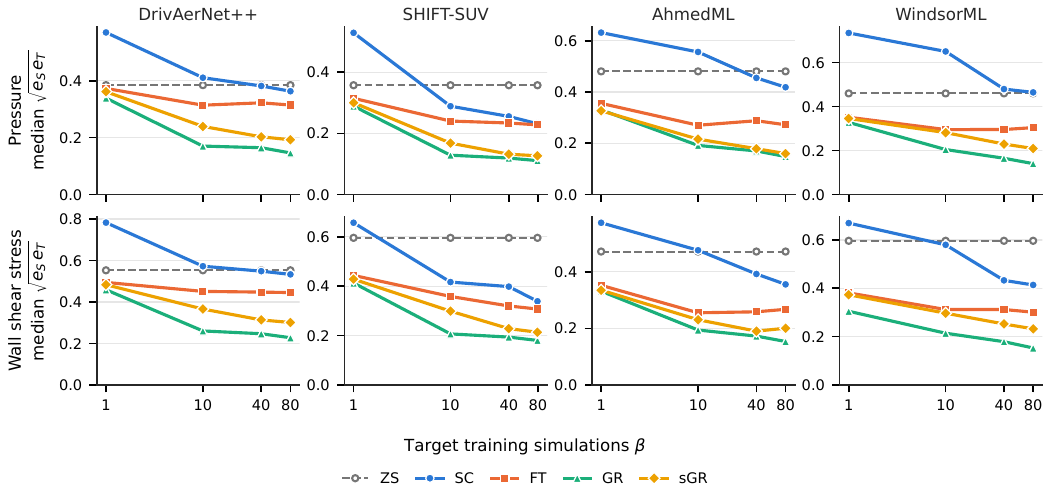}
\caption{Geometric mean $\sqrt{e_S\,e_T}$ as a function of the target training budget $\Beta$, for pressure (top) and wall shear stress (bottom), over the same surrogates in every panel. ZS is the pretrained model without adaptation.}
\label{fig:gm-budget}
\end{figure*}

\section{Experiments}
\label{sec:experiments}

\subsection{Experimental Setup}
\label{sec:setup}

\paragraph{Datasets.}
We use five independent CFD datasets, one source dataset and four target datasets (Table~\ref{tab:datasets}). The source contains the 436 DrivAerML road-car simulations used to pretrain the released models \citep{ashton2024drivaerml}. The targets contain 500 DrivAerNet++ production-car simulations \citep{elrefaie2024drivaernet++}, 99 SHIFT-SUV quarter-scale sport utility vehicle simulations \citep{shift_suv_2025}, whose 10 shape parameters and $\approx$45M-cell meshes are documented by \citet{alkin2025abuptauto} and \citet{adams2025geotransolver}, 499 complete full-body AhmedML simulations \citep{ashton2024ahmedml}, and the 350 published WindsorML square-back simulations \citep{ashton2024windsorml}.

\paragraph{Models.}
The complete evaluation pairs DrivAerNet++, SHIFT-SUV, AhmedML, and WindsorML with four DrivAerML-pretrained surface surrogates released through PhysicsNeMo-CFD \citep{tangsali2025benchmarking}: Transolver (TSol) \citep{pmlr-v235-wu24r}, GeoTransolver (GeoT) \citep{adams2025geotransolver}, DoMINO (DoM) \citep{ranade2025domino}, and X-MeshGraphNet (XMGN) \citep{nabian2024x}, and with GINO \citep{li2023geometry}. GINO has no released DrivAerML checkpoint, so we pretrain it on the same 436-simulation DrivAerML split with its published recipe (Appendix~\ref{app:implementation}). For reference, the appendix also reports AB-UPT \citep{alkin2025ab} against training from scratch (Table~\ref{tab:mainsav-b10}). It uses its released DrivAerML checkpoint and requires the cell centers of each geometry's CFD volume mesh, so we keep it outside the benchmark counts.

\paragraph{Training and evaluation.}
Each target uses ten validation and ten test simulations, except WindsorML, for which we use the official validation and test splits of the release (35 simulations each). We report the main comparison at $\Beta=10$ target training simulations, where only a few labeled target geometries are available, and repeat it at $\Beta\in\{1,40,80\}$ (Appendix~\ref{app:georeplay-budget}). SHIFT-SUV uses $\Beta=79$ at the largest budget because only 99 surface simulations are available. Each model trains until its validation error stops improving (early stopping with patience 30, at most 500 epochs), and we select the checkpoint with the lowest validation pressure and wall shear stress error. Fine-tuned and from-scratch runs use the same PhysicsNeMo model implementation and split. We train every comparison with seed 0 and repeat the Transolver and GeoTransolver comparisons with seeds 1 and 2 (Appendix~\ref{app:mean_var}). Appendix~\ref{app:implementation} contains the training recipes and dataset conversions.

For pressure and wall shear stress, we compute relative $L_2$ error on each complete surface and average it over the target test simulations for $e_T$ and over the 48 designs of the official DrivAerML validation split for $e_S$. The wall shear stress norm includes all three vector components. We integrate the predicted pressure and wall shear stress over the CFD surface to obtain drag and report mean relative drag error. Because each error is relative to the magnitude of the CFD field or force, it is dimensionless and comparable across datasets with different units and scales.

\subsection{Results}
\label{sec:results}

\paragraph{\GeoReplay{} preserves source accuracy at matched target accuracy.}
We report pressure errors at $\Beta=10$ in Table~\ref{tab:main-b10}. Fine-tuning lowers the target error of every surrogate but raises its DrivAerML error far above zero-shot prediction, from $0.042$ to $0.350$ for Transolver adapted to DrivAerNet++. \GeoReplay{} keeps the target error of fine-tuning within $0.01$ in 19 of 20 surrogate and target pairs while lowering the DrivAerML error by a median factor of $3.3$, to $0.096$ in the same example, and it has the lowest geometric mean in 18 of the 20 rows. Across all 80 combinations of surrogate, target dataset, and budget (Appendix~\ref{app:georeplay-budget}), \GeoReplay{} lowers the DrivAerML pressure error of fine-tuning in 77 and of training from scratch in all 80, while its median target error differs from fine-tuning by $0.2\%$. We report the standard deviation over three training seeds in Appendix Table~\ref{tab:main-b10-spread}. Across seeds, the standard deviation of each error is at most $9\%$ of its value and that of the geometric mean at most $3\%$, far smaller than the gap in DrivAerML error between fine-tuning and \GeoReplay{}. Among the adapted methods, \GeoReplay{} has the lowest median geometric mean for every target dataset and budget in Figure~\ref{fig:gm-budget} except AhmedML pressure at $\Beta=1$ (sparse \GeoReplay{}, $0.004$ lower). As expected, the geometric mean of every adapted method is lower at $\Beta=10$ than at $\Beta=1$. Drag shows the same pattern, with a lower drag geometric mean than fine-tuning in 19 of 20 pairs (Table~\ref{tab:drag-b10}). The pattern also holds for wall shear stress and lift, where \GeoReplay{} has a lower geometric mean than fine-tuning in 20 of 20 and 19 of 20 pairs at $\Beta=10$ (Appendix Tables~\ref{tab:georeplay-budget-wss} and~\ref{tab:georeplay-budget-lift}).

\begin{table}[!ht]
\centering
\caption{Drag geometric mean $\sqrt{e_S\,e_T}$ at $\Beta=10$ for fine-tuning (FT) and \GeoReplay{} (GR), with errors as in Table~\ref{tab:main-b10}. \textbf{Bold} marks the lower of each pair. Appendix Table~\ref{tab:georeplay-budget-drag} gives both errors and every budget. }
\label{tab:drag-b10}
\footnotesize
\setlength{\tabcolsep}{2.5pt}
\resizebox{\columnwidth}{!}{%
\begin{tabular}{llrrrrr}
\toprule
$T$ & & TSol & GeoT & DoM & XMGN & GINO \\
\midrule
\multirow{2}{*}{DrivAerNet++} & FT & 0.1314 & 0.0406 & 0.1087 & \textbf{0.1697} & 0.2065 \\
 & GR & \textbf{0.0398} & \textbf{0.0189} & \textbf{0.0287} & 0.1737 & \textbf{0.1978} \\
\midrule
\multirow{2}{*}{SHIFT-SUV} & FT & 0.0652 & 0.0577 & 0.1301 & 0.1293 & 0.1577 \\
 & GR & \textbf{0.0156} & \textbf{0.0134} & \textbf{0.0302} & \textbf{0.0970} & \textbf{0.1557} \\
\midrule
\multirow{2}{*}{AhmedML} & FT & 0.0662 & 0.0618 & 0.1973 & 0.3470 & 0.7981 \\
 & GR & \textbf{0.0435} & \textbf{0.0313} & \textbf{0.0686} & \textbf{0.3460} & \textbf{0.1655} \\
\midrule
\multirow{2}{*}{WindsorML} & FT & 0.0709 & 0.0713 & 0.2435 & 0.4175 & 0.7215 \\
 & GR & \textbf{0.0437} & \textbf{0.0400} & \textbf{0.0537} & \textbf{0.3668} & \textbf{0.1351} \\
\bottomrule
\end{tabular}}

\end{table}

\paragraph{Component ablation.}
In Table~\ref{tab:ablation-b10}, we separate the source weighting from the update rule for Transolver and GeoTransolver. Plain replay with uniform weights corresponds to experience replay \citep{rolnick2019experience} on the source simulations. Every replay variant lowers the DrivAerML error of fine-tuning by a factor of $3.3$ to $5.8$, but only \GeoReplay{} keeps the target error within $0.005$ of fine-tuning on every target dataset. The plain update raises the target error by up to $0.042$ and the aligned update with uniform weights by up to $0.022$, both on SHIFT-SUV, so both components are needed to preserve target accuracy.

\paragraph{Fine-tuning alone often loses to training from scratch.}
Without replay, fine-tuning has a lower pressure or wall shear stress error than training from scratch in only 17 of 40 comparisons at $\Beta=10$ and a higher error in 23 (Appendix~\ref{app:results}). Which arm wins depends jointly on the surrogate, target dataset, and quantity. Fine-tuned GeoTransolver has a $41\%$ higher pressure error than training from scratch on WindsorML yet stays within $2\%$ of it on DrivAerNet++, and on SHIFT-SUV fine-tuned DoMINO lowers both surface-field errors while nearly tripling the drag error. Figure~\ref{fig:qualitative} in Appendix~\ref{app:qualitative} shows that a large reduction of zero-shot error can coexist with negative transfer value.

\paragraph{Distance to the closest training geometry does not determine transfer value.}
DrivAerNet++ and SHIFT-SUV lie closer to the DrivAerML training geometries than AhmedML and WindsorML do (Figure~\ref{fig:geometry-summary}, Appendix Table~\ref{tab:geometry-distance}), and Appendix~\ref{app:geometry} (Figure~\ref{fig:test-coverage}) compares test-set coverage by the target training split and by DrivAerML. Nevertheless, fine-tuning has a lower pressure, wall shear stress, or drag error than training from scratch in only 5 of 15 surrogate and quantity comparisons on SHIFT-SUV, as rarely as on the more distant AhmedML (5 of 15) and WindsorML (4 of 15), against 9 of 15 on DrivAerNet++. Neither geometric distance nor zero-shot error determines transfer value, so geometric coverage alone cannot decide whether a pretrained surrogate remains useful after fine-tuning. The targets also differ from DrivAerML in CFD solver, turbulence model, scale, and Reynolds number (Table~\ref{tab:datasets}), which geometric distance does not capture.

\section{Discussion}
\label{sec:discussion}

Our proposed \GeoReplay{} addresses the failure mode exposed by our transfer learning analysis. Fine-tuning on a few target simulations overwrites much of the source accuracy of the pretrained surrogate, and replaying source simulations during adaptation prevents this without a measurable loss of target accuracy.

Several limitations bound these conclusions. We measure variation over three training seeds at $\Beta=10$ on one split per target, so small differences at other budgets require caution. Our geometry measures omit CFD solver, turbulence, operating-condition, mesh, and label differences that may affect transfer.

\section{Conclusion}
\label{sec:conclusion}

With few target simulations, fine-tuning a pretrained surrogate beats training from scratch in fewer than half of the comparisons and erodes its source accuracy. \GeoReplay{} lowers the source error of fine-tuning in 77 of 80 settings, by a median factor of $2.8$, while matching its target accuracy. Because geometric distance does not determine transfer value, evaluations should report both the from-scratch comparison and the source error. All released surrogates learn from DrivAerML, which is why we study this setting; pretraining across datasets is a natural next step, and Appendix~\ref{app:pretraining} reports a first comparison.

\subsection*{AI use statement}

We used generative AI tools to assist with programming, literature search, and editing the manuscript text. The authors reviewed and tested all AI-assisted code, verified all cited sources, and checked all reported results and text. We take responsibility for the final content of this work, including text, claims, and artifacts produced with the aid of generative AI.

\bibliographystyle{plainnat}
\bibliography{references}

@article{ashton2024drivaerml,
  title={{DrivAerML}: High-fidelity computational fluid dynamics dataset for road-car external aerodynamics},
  author={Ashton, Neil and Mockett, Charles and Fuchs, Marian and Fliessbach, Louis and Hetmann, Hendrik and Knacke, Thilo and Schonwald, Norbert and Skaperdas, Vangelis and Fotiadis, Grigoris and Walle, Astrid and others},
  journal={arXiv preprint arXiv:2408.11969},
  year={2024}
}

@article{ashton2024windsorml,
  title={{WindsorML}: High-fidelity computational fluid dynamics dataset for automotive aerodynamics},
  author={Ashton, Neil and Angel, Jordan B and Ghate, Aditya S and Kenway, Gaetan K. W. and Wong, Man Long and Kiris, Cetin and Walle, Astrid and Maddix, Danielle C. and Page, Gary},
  journal={arXiv preprint arXiv:2407.19320},
  year={2024}
}

@inproceedings{ananthan2024machine,
  title={Machine learning for road vehicle aerodynamics},
  author={Ananthan, Vidyasagar and Ashton, Neil and Chadwick, Nate and Lizarraga, Mariano and Maddix, Danielle C. and Maheswaran, Satheesh and Moreno, Pablo Hermoso and M. Shabestari, Parisa and Sovani, Sandeep and Subramanian, Shreyas and others},
  booktitle={SAE Technical Paper 2024-01-2529},
  year={2024}
}

@article{elrefaie2024drivaernet++,
  title={{Drivaernet++}: A large-scale multimodal car dataset with computational fluid dynamics simulations and deep learning benchmarks},
  author={Elrefaie, Mohamed and Morar, Florin and Dai, Angela and Ahmed, Faez},
  journal={Advances in Neural Information Processing Systems},
  volume={37},
  pages={499--536},
  year={2024}
}

@misc{shift_suv_2025,
  howpublished = {Hugging Face dataset luminary-shift/SUV-sample},
  author = {{Luminary Cloud}},
  title  = {{SHIFT-SUV Sample}: High-Fidelity Computational Fluid Dynamics Dataset for {SUV} External Aerodynamics},
  year   = {2025},
}

@article{alkin2025ab,
  title={{AB-UPT}: Scaling neural {CFD} surrogates for high-fidelity automotive aerodynamics simulations via anchored-branched universal physics transformers},
  author={Alkin, Benedikt and Bleeker, Maurits and Kurle, Richard and Kronlachner, Tobias and Sonnleitner, Reinhard and Dorfer, Matthias and Brandstetter, Johannes},
  journal={Transactions on Machine Learning Research},
  year={2025}
}

@article{adams2025geotransolver,
  title={{GeoTransolver}: Learning physics on irregular domains using multi-scale geometry aware physics attention transformer},
  author={Adams, Corey and Ranade, Rishikesh and Cherukuri, Ram and Choudhry, Sanjay},
  journal={arXiv preprint arXiv:2512.20399},
  year={2025}
}

@article{ranade2025domino,
  title={{DoMINO}: A decomposable multi-scale iterative neural operator for modeling large scale engineering simulations},
  author={Ranade, Rishikesh and Nabian, Mohammad Amin and Tangsali, Kaustubh and Kamenev, Alexey and Hennigh, Oliver and Cherukuri, Ram and Choudhry, Sanjay},
  journal={arXiv preprint arXiv:2501.13350},
  year={2025}
}

@article{choy2025factorized,
  title={Factorized implicit global convolution for automotive computational fluid dynamics prediction},
  author={Choy, Chris and Kamenev, Alexey and Kossaifi, Jean and Rietmann, Max and Kautz, Jan and Azizzadenesheli, Kamyar},
  journal={arXiv preprint arXiv:2502.04317},
  year={2025}
}

@article{nabian2024x,
  title={{X-MeshGraphNet}: Scalable multi-scale graph neural networks for physics simulation},
  author={Nabian, Mohammad Amin and Liu, Chang and Ranade, Rishikesh and Choudhry, Sanjay},
  journal={arXiv preprint arXiv:2411.17164},
  year={2024}
}

@InProceedings{pmlr-v235-wu24r,
  title = 	 {Transolver: A Fast Transformer Solver for {PDE}s on General Geometries},
  author =       {Wu, Haixu and Luo, Huakun and Wang, Haowen and Wang, Jianmin and Long, Mingsheng},
  booktitle = 	 {Proceedings of the 41st International Conference on Machine Learning},
  pages = 	 {53681--53705},
  year = 	 {2024},
  editor = 	 {Salakhutdinov, Ruslan and Kolter, Zico and Heller, Katherine and Weller, Adrian and Oliver, Nuria and Scarlett, Jonathan and Berkenkamp, Felix},
  volume = 	 {235},
  series = 	 {Proceedings of Machine Learning Research},
  month = 	 {21--27 Jul},
  publisher =    {PMLR},
}

@article{tangsali2025benchmarking,
  title={A benchmarking framework for {AI} models in automotive aerodynamics},
  author={Tangsali, Kaustubh and Ranade, Rishikesh and Nabian, Mohammad Amin and Kamenev, Alexey and Sharpe, Peter and Ashton, Neil and Cherukuri, Ram and Choudhry, Sanjay},
  journal={arXiv preprint arXiv:2507.10747},
  year={2025}
}

@article{elrefaie2025carbench,
  title={{CarBench}: A comprehensive benchmark for neural surrogates on high-fidelity {3D} car aerodynamics},
  author={Elrefaie, Mohamed and Shu, Dule and Klenk, Matt and Ahmed, Faez},
  journal={arXiv preprint arXiv:2512.07847},
  year={2025}
}

@article{breakevencomplexity,
  title={Breakeven complexity: A new perspective on neural partial differential equation solvers},
  author={Zhang, Yijing and Roberts, Nicholas and Marwah, Tanya and Khodak, Mikhail},
  journal={arXiv preprint arXiv:2605.15399},
  year={2026}
}

@misc{physicsxlpm,
  title  = {Scaling Physics {AI} for Automotive Aerodynamics},
  author = {{PhysicsX Research}},
  year   = {2026},
  month  = mar,
  note   = {Industry technical report},
}

@misc{nvidia,
  howpublished = {NVIDIA Blog},
  title  = {{NVIDIA AI} Physics Transforms Aerospace and Automotive Design, Accelerating Engineering by 500x},
  author = {Costa, Timothy},
  year   = {2025},
  month  = oct,
}

@article{subramanian2023towards,
  title={Towards foundation models for scientific machine learning: Characterizing scaling and transfer behavior},
  author={Subramanian, Shashank and Harrington, Peter and Keutzer, Kurt and Bhimji, Wahid and Morozov, Dmitriy and Mahoney, Michael W and Gholami, Amir},
  journal={Advances in neural information processing systems},
  volume={36},
  pages={71242--71262},
  year={2023}
}

@inproceedings{multiple_physics,
  title     = {Multiple Physics Pretraining for Spatiotemporal Surrogate Models},
  author    = {Michael McCabe and Bruno R{\'e}galdo-Saint Blancard and Liam Parker and Ruben Ohana and Miles Cranmer and Alberto Bietti and Michael Eickenberg and Siavash Golkar and Geraud Krawezik and Francois Lanusse and Mariel Pettee and Tiberiu Tesileanu and Kyunghyun Cho and Shirley Ho},
  booktitle = {Advances in Neural Information Processing Systems},
  volume    = {37},
  year      = {2024},
}

@inproceedings{mouli2024using,
  title={Using Uncertainty Quantification to Characterize and Improve Out-of-Domain Learning for {PDEs}},
  author={Mouli, S Chandra and Maddix, Danielle C. and Alizadeh, Shima and Gupta, Gaurav and Stuart, Andrew and Mahoney, Michael W and Wang, Yuyang},
  booktitle={Proceedings of the 41st International Conference on Machine Learning},
  series={Proceedings of Machine Learning Research},
  volume={235},
  year={2024}
}

@InProceedings{pmlr-v235-berman24b,
  title = 	 {{C}o{L}o{RA}: Continuous low-rank adaptation for reduced implicit neural modeling of parameterized partial differential equations},
  author =       {Berman, Jules and Peherstorfer, Benjamin},
  booktitle = 	 {Proceedings of the 41st International Conference on Machine Learning},
  pages = 	 {3565--3583},
  year = 	 {2024},
  editor = 	 {Salakhutdinov, Ruslan and Kolter, Zico and Heller, Katherine and Weller, Adrian and Oliver, Nuria and Scarlett, Jonathan and Berkenkamp, Felix},
  volume = 	 {235},
  series = 	 {Proceedings of Machine Learning Research},
  month = 	 {21--27 Jul},
  publisher =    {PMLR}
}

@inproceedings{cao2025spectral,
  title={{Spectral-Refiner}: Accurate fine-tuning of spatiotemporal {Fourier} neural operator for turbulent flows},
  author={Cao, Shuhao and Brarda, Francesco and Li, Ruipeng and Xi, Yuanzhe},
  booktitle={International Conference on Learning Representations},
  pages={60348--60384},
  year={2025}
}

@inproceedings{wang2019characterizing,
  title={Characterizing and avoiding negative transfer},
  author={Wang, Zirui and Dai, Zihang and P{\'o}czos, Barnab{\'a}s and Carbonell, Jaime},
  booktitle={2019 IEEE/CVF Conference on Computer Vision and Pattern Recognition (CVPR)},
  pages={11285--11294},
  year={2019},
  organization={IEEE}
}

@article{hanneke2019value,
  title={On the value of target data in transfer learning},
  author={Hanneke, Steve and Kpotufe, Samory},
  journal={Advances in Neural Information Processing Systems},
  volume={32},
  year={2019}
}

@inproceedings{regression_transferability,
  title     = {Simple Transferability Estimation for Regression Tasks},
  author    = {Cuong N. Nguyen and Phong Tran and Lam Si Tung Ho and Vu Dinh and Anh T. Tran and Tal Hassner and Cuong V. Nguyen},
  booktitle = {Proceedings of the Thirty-Ninth Conference on Uncertainty in Artificial Intelligence},
  series    = {Proceedings of Machine Learning Research},
  volume    = {216},
  pages     = {1510--1521},
  year      = {2023},
  publisher = {PMLR},
}

@inproceedings{nguyen2020leep,
  title={Leep: A new measure to evaluate transferability of learned representations},
  author={Nguyen, Cuong and Hassner, Tal and Seeger, Matthias and Archambeau, Cedric},
  booktitle={International conference on machine learning},
  pages={7294--7305},
  year={2020},
  organization={PMLR}
}

@article{ashton2024ahmedml,
  title={{AhmedML}: High-fidelity computational fluid dynamics dataset for incompressible, low-speed bluff body aerodynamics},
  author={Ashton, Neil and Maddix, Danielle C. and Gundry, Samuel and Shabestari, Parisa M},
  journal={arXiv preprint arXiv:2407.20801},
  year={2024}
}

@inproceedings{cuturi2013sinkhorn,
  title     = {Sinkhorn Distances: Lightspeed Computation of Optimal Transport},
  author    = {Cuturi, Marco},
  booktitle = {Advances in Neural Information Processing Systems},
  volume    = {26},
  year      = {2013},
}

@inproceedings{yu2020gradient,
  title     = {Gradient Surgery for Multi-Task Learning},
  author    = {Yu, Tianhe and Kumar, Saurabh and Gupta, Abhishek and Levine, Sergey and Hausman, Karol and Finn, Chelsea},
  booktitle = {Advances in Neural Information Processing Systems},
  volume    = {33},
  year      = {2020},
}

@inproceedings{li2023geometry,
  title     = {Geometry-Informed Neural Operator for Large-Scale {3D} {PDEs}},
  author    = {Li, Zongyi and Kovachki, Nikola Borislavov and Choy, Chris and Li, Boyi and Kossaifi, Jean and Otta, Shourya Prakash and Nabian, Mohammad Amin and Stadler, Maximilian and Hundt, Christian and Azizzadenesheli, Kamyar and Anandkumar, Anima},
  booktitle = {Advances in Neural Information Processing Systems},
  volume    = {36},
  year      = {2023}
}

@article{alkin2025abuptauto,
  title={{AB-UPT} for Automotive and Aerospace Applications},
  author={Alkin, Benedikt and Kurle, Richard and Serrano, Louis and Just, Dennis and Brandstetter, Johannes},
  journal={arXiv preprint arXiv:2510.15808},
  year={2025}
}

@inproceedings{li2021fourier,
  title={Fourier Neural Operator for Parametric Partial Differential Equations},
  author={Li, Zongyi and Kovachki, Nikola and Azizzadenesheli, Kamyar and Liu, Burigede and Bhattacharya, Kaushik and Stuart, Andrew and Anandkumar, Anima},
  booktitle={International Conference on Learning Representations},
  year={2021}
}

@article{bommasani2021opportunities,
  title={On the Opportunities and Risks of Foundation Models},
  author={Bommasani, Rishi and Hudson, Drew A. and Adeli, Ehsan and Altman, Russ and Arora, Simran and von Arx, Sydney and Bernstein, Michael S. and Bohg, Jeannette and Bosselut, Antoine and Brunskill, Emma and others},
  journal={arXiv preprint arXiv:2108.07258},
  year={2021}
}

@article{du2018adapting,
  title={Adapting Auxiliary Losses Using Gradient Similarity},
  author={Du, Yunshu and Czarnecki, Wojciech M. and Jayakumar, Siddhant M. and Farajtabar, Mehrdad and Pascanu, Razvan and Lakshminarayanan, Balaji},
  journal={arXiv preprint arXiv:1812.02224},
  year={2018}
}

@inproceedings{lopezpaz2017gem,
  title={Gradient Episodic Memory for Continual Learning},
  author={Lopez-Paz, David and Ranzato, Marc'Aurelio},
  booktitle={Advances in Neural Information Processing Systems},
  volume={30},
  year={2017}
}

@inproceedings{chaudhry2019agem,
  title={Efficient Lifelong Learning with {A-GEM}},
  author={Chaudhry, Arslan and Ranzato, Marc'Aurelio and Rohrbach, Marcus and Elhoseiny, Mohamed},
  booktitle={International Conference on Learning Representations},
  year={2019}
}

@inproceedings{rolnick2019experience,
  title={Experience Replay for Continual Learning},
  author={Rolnick, David and Ahuja, Arun and Schwarz, Jonathan and Lillicrap, Timothy and Wayne, Gregory},
  booktitle={Advances in Neural Information Processing Systems},
  volume={32},
  year={2019}
}

@article{kirkpatrick2017overcoming,
  title={Overcoming Catastrophic Forgetting in Neural Networks},
  author={Kirkpatrick, James and Pascanu, Razvan and Rabinowitz, Neil and Veness, Joel and Desjardins, Guillaume and Rusu, Andrei A. and Milan, Kieran and Quan, John and Ramalho, Tiago and Grabska-Barwinska, Agnieszka and others},
  journal={Proceedings of the National Academy of Sciences},
  volume={114},
  number={13},
  pages={3521--3526},
  year={2017}
}

@inproceedings{li2018explicit,
  title={Explicit Inductive Bias for Transfer Learning with Convolutional Networks},
  author={Li, Xuhong and Grandvalet, Yves and Davoine, Franck},
  booktitle={International Conference on Machine Learning},
  pages={2825--2834},
  year={2018}
}

@inproceedings{courty2017joint,
  title={Joint Distribution Optimal Transportation for Domain Adaptation},
  author={Courty, Nicolas and Flamary, R{\'e}mi and Habrard, Amaury and Rakotomamonjy, Alain},
  booktitle={Advances in Neural Information Processing Systems},
  volume={30},
  pages={3730--3739},
  year={2017}
}

@inproceedings{redko2017theoretical,
  title={Theoretical Analysis of Domain Adaptation with Optimal Transport},
  author={Redko, Ievgen and Habrard, Amaury and Sebban, Marc},
  booktitle={Joint European Conference on Machine Learning and Knowledge Discovery in Databases (ECML PKDD)},
  year={2017},
  pages={737--753}
}

@inproceedings{shen2018wasserstein,
  title={Wasserstein Distance Guided Representation Learning for Domain Adaptation},
  author={Shen, Jian and Qu, Yanru and Zhang, Weinan and Yu, Yong},
  booktitle={Proceedings of the AAAI Conference on Artificial Intelligence},
  volume={32},
  pages={4058--4065},
  year={2018}
}

@inproceedings{herde2024poseidon,
  title={Poseidon: Efficient Foundation Models for {PDEs}},
  author={Herde, Maximilian and Raoni{\'c}, Bogdan and Rohner, Tobias and K{\"a}ppeli, Roger and Molinaro, Roberto and de B{\'e}zenac, Emmanuel and Mishra, Siddhartha},
  booktitle={Advances in Neural Information Processing Systems},
  volume={37},
  year={2024}
}

@article{hagnberger2026smart,
  title={{SMART}: Scalable Mesh-free Aerodynamic Simulations from Raw Geometries using a Transformer-based Surrogate Model},
  author={Hagnberger, Jan and Niepert, Mathias},
  journal={arXiv preprint arXiv:2601.18707},
  year={2026}
}

@article{zhou2026crossdomain,
  title={Cross-Domain Pretraining for Steady-State Neural {CFD} Surrogates},
  author={Zhou, Anthony and Barati Farimani, Amir and Ho, Shirley and Morel, Rudy},
  journal={arXiv preprint arXiv:2610.10398},
  year={2026}
}

\clearpage
\appendix
\thispagestyle{empty}
\onecolumn
\aistatstitle{Measuring and Improving Transfer of Neural Surrogates for Vehicle Aerodynamics: Appendix}

\FloatBarrier
\section{Geometry Analysis}
\label{app:geometry}

As a within-source calibration, we compare the official 48-geometry DrivAerML validation split with the same 436 pretraining geometries. Its median closest-training distance is $0.0183$ body lengths, with an interquartile range of $[0.0179,\,0.0189]$.

\begin{figure*}[!ht]
\centering
\begin{minipage}[t]{0.55\textwidth}
\vspace{0pt}\centering
\includegraphics[width=\linewidth]{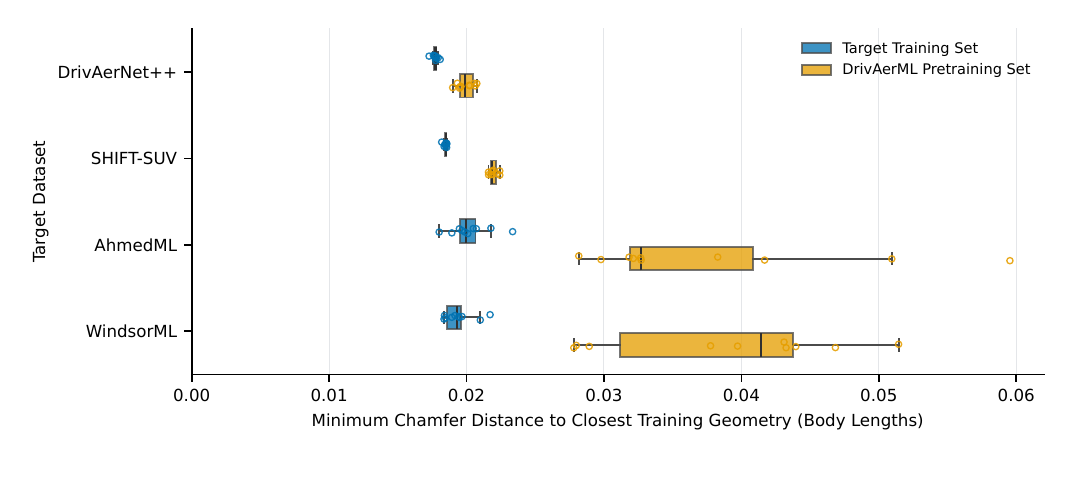}
\caption{Minimum Chamfer distance from each held-out target geometry to the target training set (blue) or DrivAerML pretraining set (orange). Within each target, both distributions contain the same test geometries; boxes show the median and interquartile range, and whiskers extend to 1.5 times the interquartile range.}
\label{fig:test-coverage}
\end{minipage}\hfill
\begin{minipage}[t]{0.42\textwidth}
\vspace{0pt}\centering
\captionof{table}{Nearest-source geometry distance for all usable geometries in four target datasets, measured against the closest of 436 DrivAerML pretraining geometries. Values are in body lengths; lower is closer. \textbf{Bold}/\underline{underlined} mark the lowest/second-lowest.}
\label{tab:geometry-distance}
\small
\setlength{\tabcolsep}{3pt}
\begin{tabular}{lrcc}
\toprule
Target dataset & $N$ & $\mathrm{Median}_{[\mathrm{IQR}]}\downarrow$ & 95th pct.$\downarrow$ \\
\midrule
DrivAerNet++ & 500 & $\mathbf{0.020}_{[0.020,\,0.020]}$ & 0.021 \\
SHIFT-SUV & 99 & $\underline{0.022}_{[0.022,\,0.022]}$ & 0.022 \\
AhmedML & 499 & $0.036_{[0.031,\,0.043]}$ & 0.061 \\
WindsorML & 350 & $0.034_{[0.030,\,0.043]}$ & 0.052 \\
\bottomrule
\end{tabular}
\end{minipage}
\end{figure*}

We sample 1,024 area-weighted surface points per geometry and compute symmetric Chamfer distance by averaging the target-to-source and source-to-target nearest-point distances. We reconcile dataset axes and apply uniform body-length normalization before sampling, which produces a dimensionless distance while retaining differences in vehicle proportions. For Figure~\ref{fig:geometry-summary}, we compare every usable geometry from the four target datasets with all 436 DrivAerML training geometries. For the within-dataset analysis, we form all unordered pairs within the 436 DrivAerML, 500 DrivAerNet++, 99 SHIFT-SUV and 350 WindsorML geometries. Table~\ref{tab:geometry-distance} retains the closest DrivAerML geometry for every usable target geometry, including the 499-geometry AhmedML extension. Figure~\ref{fig:test-coverage} instead uses the ten fixed test geometries per complete target and compares their nearest DrivAerML and target-training geometries. As robustness checks, we also evaluate sliced Wasserstein distance, using 64 projection directions and 64 quantiles per direction, and a robust-standardized descriptor containing width-to-length and height-to-length ratios, vertical quantiles, and coarse longitudinal body profiles.

\FloatBarrier
\clearpage
\section{Variance Across Different Seeds}
\label{app:mean_var}

\par\noindent\begin{minipage}{\textwidth}%
\centering
\captionof{table}{Table~\ref{tab:main-b10} transposed, with the standard deviation over three training seeds. Columns are surrogates, and rows give the source error $e_S$, the target error $e_T$, and their geometric mean (shaded) for each method and target dataset $T$. Values are those of Table~\ref{tab:main-b10} (mean) with subscripts of the standard deviation over all seeds. Within each surrogate, target dataset, and error, \textbf{bold} marks the lowest of the five methods.}
\label{tab:main-b10-spread}
\fontsize{8}{9}\selectfont
\setlength{\tabcolsep}{7pt}
\renewcommand{\arraystretch}{0.9}
\begin{tabular}{lllrrrrr}
\toprule
$T$ & Error & Method & TSol $\downarrow$ & GeoT $\downarrow$ & DoM $\downarrow$ & XMGN $\downarrow$ & GINO $\downarrow$ \\
\midrule
\multirow{15}{*}{\rotatebox[origin=c]{90}{DrivAerNet++}} & $e_S$ & ZS & \textbf{0.042}$_{\pm0.002}$ & \textbf{0.037}$_{\pm0.002}$ & 0.186$_{\pm0.009}$ & \textbf{0.278}$_{\pm0.009}$ & \textbf{0.268}$_{\pm0.003}$ \\
 &  & SC & 0.736$_{\pm0.005}$ & 0.764$_{\pm0.002}$ & 0.830$_{\pm0.004}$ & 0.719$_{\pm0.003}$ & 0.777$_{\pm0.017}$ \\
 &  & FT & 0.350$_{\pm0.002}$ & 0.199$_{\pm0.005}$ & 0.461$_{\pm0.006}$ & 0.499$_{\pm0.010}$ & 0.555$_{\pm0.005}$ \\
 &  & GR & 0.096$_{\pm0.004}$ & 0.060$_{\pm0.000}$ & \textbf{0.135}$_{\pm0.004}$ & 0.392$_{\pm0.009}$ & 0.360$_{\pm0.007}$ \\
 &  & sGR & 0.290$_{\pm0.003}$ & 0.063$_{\pm0.001}$ & 0.232$_{\pm0.006}$ & 0.371$_{\pm0.008}$ & 0.432$_{\pm0.007}$ \\
\cmidrule(l){2-8}
 & $e_T$ & ZS & 1.020$_{\pm0.030}$ & 1.271$_{\pm0.037}$ & 0.794$_{\pm0.042}$ & 1.194$_{\pm0.055}$ & 0.865$_{\pm0.040}$ \\
 &  & SC & 0.215$_{\pm0.003}$ & \textbf{0.202}$_{\pm0.005}$ & 0.242$_{\pm0.014}$ & 0.234$_{\pm0.018}$ & \textbf{0.283}$_{\pm0.019}$ \\
 &  & FT & \textbf{0.197}$_{\pm0.000}$ & 0.205$_{\pm0.002}$ & \textbf{0.214}$_{\pm0.013}$ & 0.226$_{\pm0.013}$ & \textbf{0.283}$_{\pm0.012}$ \\
 &  & GR & \textbf{0.197}$_{\pm0.001}$ & 0.206$_{\pm0.002}$ & 0.215$_{\pm0.013}$ & \textbf{0.221}$_{\pm0.014}$ & 0.286$_{\pm0.012}$ \\
 &  & sGR & \textbf{0.197}$_{\pm0.010}$ & 0.208$_{\pm0.013}$ & \textbf{0.214}$_{\pm0.013}$ & \textbf{0.221}$_{\pm0.014}$ & 0.284$_{\pm0.012}$ \\
\cmidrule(l){2-8}
 & $\sqrt{e_S e_T}$ & \cellcolor{gmshade}ZS & \cellcolor{gmshade}0.207$_{\pm0.006}$ & \cellcolor{gmshade}0.217$_{\pm0.006}$ & \cellcolor{gmshade}0.384$_{\pm0.014}$ & \cellcolor{gmshade}0.576$_{\pm0.016}$ & \cellcolor{gmshade}0.481$_{\pm0.012}$ \\
 &  & \cellcolor{gmshade}SC & \cellcolor{gmshade}0.398$_{\pm0.003}$ & \cellcolor{gmshade}0.393$_{\pm0.005}$ & \cellcolor{gmshade}0.448$_{\pm0.013}$ & \cellcolor{gmshade}0.410$_{\pm0.016}$ & \cellcolor{gmshade}0.469$_{\pm0.016}$ \\
 &  & \cellcolor{gmshade}FT & \cellcolor{gmshade}0.263$_{\pm0.001}$ & \cellcolor{gmshade}0.202$_{\pm0.001}$ & \cellcolor{gmshade}0.314$_{\pm0.010}$ & \cellcolor{gmshade}0.336$_{\pm0.011}$ & \cellcolor{gmshade}0.397$_{\pm0.008}$ \\
 &  & \cellcolor{gmshade}GR & \cellcolor{gmshade}\textbf{0.137}$_{\pm0.003}$ & \cellcolor{gmshade}\textbf{0.111}$_{\pm0.001}$ & \cellcolor{gmshade}\textbf{0.170}$_{\pm0.006}$ & \cellcolor{gmshade}0.294$_{\pm0.010}$ & \cellcolor{gmshade}\textbf{0.321}$_{\pm0.007}$ \\
 &  & \cellcolor{gmshade}sGR & \cellcolor{gmshade}0.239$_{\pm0.006}$ & \cellcolor{gmshade}0.114$_{\pm0.004}$ & \cellcolor{gmshade}0.223$_{\pm0.007}$ & \cellcolor{gmshade}\textbf{0.286}$_{\pm0.010}$ & \cellcolor{gmshade}0.350$_{\pm0.008}$ \\
\midrule
\multirow{15}{*}{\rotatebox[origin=c]{90}{SHIFT-SUV}} & $e_S$ & ZS & \textbf{0.042}$_{\pm0.002}$ & \textbf{0.037}$_{\pm0.002}$ & 0.186$_{\pm0.009}$ & \textbf{0.278}$_{\pm0.009}$ & \textbf{0.268}$_{\pm0.003}$ \\
 &  & SC & 0.722$_{\pm0.012}$ & 0.739$_{\pm0.005}$ & 1.261$_{\pm0.030}$ & 0.814$_{\pm0.023}$ & 0.720$_{\pm0.011}$ \\
 &  & FT & 0.414$_{\pm0.002}$ & 0.236$_{\pm0.005}$ & 0.485$_{\pm0.005}$ & 0.714$_{\pm0.022}$ & 0.972$_{\pm0.022}$ \\
 &  & GR & 0.080$_{\pm0.001}$ & 0.057$_{\pm0.000}$ & \textbf{0.134}$_{\pm0.004}$ & 0.445$_{\pm0.014}$ & 0.428$_{\pm0.010}$ \\
 &  & sGR & 0.290$_{\pm0.004}$ & 0.057$_{\pm0.001}$ & 0.167$_{\pm0.005}$ & 0.396$_{\pm0.012}$ & 0.509$_{\pm0.013}$ \\
\cmidrule(l){2-8}
 & $e_T$ & ZS & 0.733$_{\pm0.005}$ & 0.845$_{\pm0.005}$ & 0.687$_{\pm0.003}$ & 1.110$_{\pm0.015}$ & 0.582$_{\pm0.026}$ \\
 &  & SC & 0.103$_{\pm0.003}$ & \textbf{0.067}$_{\pm0.001}$ & 0.134$_{\pm0.007}$ & \textbf{0.103}$_{\pm0.009}$ & \textbf{0.221}$_{\pm0.012}$ \\
 &  & FT & \textbf{0.097}$_{\pm0.000}$ & 0.081$_{\pm0.000}$ & \textbf{0.119}$_{\pm0.007}$ & 0.117$_{\pm0.009}$ & 0.239$_{\pm0.007}$ \\
 &  & GR & 0.098$_{\pm0.000}$ & 0.082$_{\pm0.001}$ & 0.124$_{\pm0.007}$ & 0.111$_{\pm0.009}$ & 0.239$_{\pm0.007}$ \\
 &  & sGR & 0.098$_{\pm0.009}$ & 0.082$_{\pm0.009}$ & 0.120$_{\pm0.007}$ & 0.111$_{\pm0.009}$ & 0.239$_{\pm0.007}$ \\
\cmidrule(l){2-8}
 & $\sqrt{e_S e_T}$ & \cellcolor{gmshade}ZS & \cellcolor{gmshade}0.175$_{\pm0.004}$ & \cellcolor{gmshade}0.177$_{\pm0.004}$ & \cellcolor{gmshade}0.358$_{\pm0.008}$ & \cellcolor{gmshade}0.556$_{\pm0.009}$ & \cellcolor{gmshade}0.395$_{\pm0.009}$ \\
 &  & \cellcolor{gmshade}SC & \cellcolor{gmshade}0.272$_{\pm0.006}$ & \cellcolor{gmshade}0.223$_{\pm0.002}$ & \cellcolor{gmshade}0.411$_{\pm0.011}$ & \cellcolor{gmshade}0.289$_{\pm0.014}$ & \cellcolor{gmshade}0.399$_{\pm0.011}$ \\
 &  & \cellcolor{gmshade}FT & \cellcolor{gmshade}0.200$_{\pm0.001}$ & \cellcolor{gmshade}0.138$_{\pm0.002}$ & \cellcolor{gmshade}0.240$_{\pm0.007}$ & \cellcolor{gmshade}0.289$_{\pm0.011}$ & \cellcolor{gmshade}0.482$_{\pm0.009}$ \\
 &  & \cellcolor{gmshade}GR & \cellcolor{gmshade}\textbf{0.089}$_{\pm0.000}$ & \cellcolor{gmshade}\textbf{0.068}$_{\pm0.000}$ & \cellcolor{gmshade}\textbf{0.129}$_{\pm0.004}$ & \cellcolor{gmshade}0.222$_{\pm0.009}$ & \cellcolor{gmshade}\textbf{0.320}$_{\pm0.006}$ \\
 &  & \cellcolor{gmshade}sGR & \cellcolor{gmshade}0.168$_{\pm0.008}$ & \cellcolor{gmshade}0.069$_{\pm0.004}$ & \cellcolor{gmshade}0.141$_{\pm0.005}$ & \cellcolor{gmshade}\textbf{0.210}$_{\pm0.009}$ & \cellcolor{gmshade}0.349$_{\pm0.007}$ \\
\midrule
\multirow{15}{*}{\rotatebox[origin=c]{90}{AhmedML}} & $e_S$ & ZS & \textbf{0.042}$_{\pm0.002}$ & \textbf{0.037}$_{\pm0.002}$ & 0.186$_{\pm0.009}$ & \textbf{0.278}$_{\pm0.009}$ & \textbf{0.268}$_{\pm0.003}$ \\
 &  & SC & 0.824$_{\pm0.029}$ & 0.809$_{\pm0.008}$ & 1.306$_{\pm0.032}$ & 0.941$_{\pm0.011}$ & 1.707$_{\pm0.040}$ \\
 &  & FT & 0.272$_{\pm0.022}$ & 0.226$_{\pm0.020}$ & 0.286$_{\pm0.007}$ & 0.396$_{\pm0.008}$ & 0.579$_{\pm0.009}$ \\
 &  & GR & 0.075$_{\pm0.000}$ & 0.058$_{\pm0.000}$ & \textbf{0.146}$_{\pm0.005}$ & 0.332$_{\pm0.006}$ & 0.270$_{\pm0.004}$ \\
 &  & sGR & 0.211$_{\pm0.003}$ & 0.066$_{\pm0.002}$ & 0.184$_{\pm0.007}$ & 0.352$_{\pm0.007}$ & 0.277$_{\pm0.004}$ \\
\cmidrule(l){2-8}
 & $e_T$ & ZS & 1.041$_{\pm0.035}$ & 0.988$_{\pm0.033}$ & 1.242$_{\pm0.054}$ & 2.036$_{\pm0.068}$ & 1.043$_{\pm0.045}$ \\
 &  & SC & \textbf{0.136}$_{\pm0.005}$ & \textbf{0.124}$_{\pm0.003}$ & \textbf{0.243}$_{\pm0.051}$ & 0.329$_{\pm0.049}$ & 0.590$_{\pm0.085}$ \\
 &  & FT & 0.165$_{\pm0.014}$ & 0.148$_{\pm0.009}$ & 0.256$_{\pm0.050}$ & 0.292$_{\pm0.051}$ & 0.486$_{\pm0.097}$ \\
 &  & GR & 0.170$_{\pm0.008}$ & 0.149$_{\pm0.004}$ & 0.253$_{\pm0.048}$ & \textbf{0.288}$_{\pm0.052}$ & 0.474$_{\pm0.094}$ \\
 &  & sGR & 0.165$_{\pm0.052}$ & 0.145$_{\pm0.056}$ & 0.254$_{\pm0.048}$ & 0.297$_{\pm0.047}$ & \textbf{0.470}$_{\pm0.092}$ \\
\cmidrule(l){2-8}
 & $\sqrt{e_S e_T}$ & \cellcolor{gmshade}ZS & \cellcolor{gmshade}0.209$_{\pm0.006}$ & \cellcolor{gmshade}0.192$_{\pm0.005}$ & \cellcolor{gmshade}0.481$_{\pm0.015}$ & \cellcolor{gmshade}0.753$_{\pm0.017}$ & \cellcolor{gmshade}0.528$_{\pm0.012}$ \\
 &  & \cellcolor{gmshade}SC & \cellcolor{gmshade}0.335$_{\pm0.007}$ & \cellcolor{gmshade}0.317$_{\pm0.006}$ & \cellcolor{gmshade}0.563$_{\pm0.061}$ & \cellcolor{gmshade}0.557$_{\pm0.041}$ & \cellcolor{gmshade}1.003$_{\pm0.073}$ \\
 &  & \cellcolor{gmshade}FT & \cellcolor{gmshade}0.212$_{\pm0.003}$ & \cellcolor{gmshade}0.182$_{\pm0.004}$ & \cellcolor{gmshade}0.271$_{\pm0.027}$ & \cellcolor{gmshade}0.340$_{\pm0.030}$ & \cellcolor{gmshade}0.530$_{\pm0.054}$ \\
 &  & \cellcolor{gmshade}GR & \cellcolor{gmshade}\textbf{0.113}$_{\pm0.002}$ & \cellcolor{gmshade}\textbf{0.093}$_{\pm0.002}$ & \cellcolor{gmshade}\textbf{0.192}$_{\pm0.019}$ & \cellcolor{gmshade}\textbf{0.309}$_{\pm0.028}$ & \cellcolor{gmshade}\textbf{0.358}$_{\pm0.035}$ \\
 &  & \cellcolor{gmshade}sGR & \cellcolor{gmshade}0.187$_{\pm0.029}$ & \cellcolor{gmshade}0.098$_{\pm0.019}$ & \cellcolor{gmshade}0.216$_{\pm0.021}$ & \cellcolor{gmshade}0.323$_{\pm0.026}$ & \cellcolor{gmshade}0.361$_{\pm0.035}$ \\
\midrule
\multirow{15}{*}{\rotatebox[origin=c]{90}{WindsorML}} & $e_S$ & ZS & \textbf{0.042}$_{\pm0.002}$ & \textbf{0.037}$_{\pm0.002}$ & 0.186$_{\pm0.009}$ & \textbf{0.278}$_{\pm0.009}$ & \textbf{0.268}$_{\pm0.003}$ \\
 &  & SC & 1.227$_{\pm0.250}$ & 0.805$_{\pm0.017}$ & 1.808$_{\pm0.049}$ & 0.870$_{\pm0.008}$ & 0.901$_{\pm0.008}$ \\
 &  & FT & 0.333$_{\pm0.032}$ & 0.223$_{\pm0.033}$ & 0.296$_{\pm0.006}$ & 0.451$_{\pm0.008}$ & 0.867$_{\pm0.012}$ \\
 &  & GR & 0.076$_{\pm0.001}$ & 0.058$_{\pm0.001}$ & \textbf{0.144}$_{\pm0.006}$ & 0.348$_{\pm0.009}$ & 0.271$_{\pm0.004}$ \\
 &  & sGR & 0.310$_{\pm0.007}$ & 0.061$_{\pm0.001}$ & 0.169$_{\pm0.007}$ & 0.377$_{\pm0.012}$ & 0.280$_{\pm0.004}$ \\
\cmidrule(l){2-8}
 & $e_T$ & ZS & 0.794$_{\pm0.017}$ & 0.744$_{\pm0.011}$ & 1.519$_{\pm0.028}$ & 0.900$_{\pm0.012}$ & 0.790$_{\pm0.013}$ \\
 &  & SC & 0.427$_{\pm0.113}$ & \textbf{0.173}$_{\pm0.005}$ & \textbf{0.279}$_{\pm0.018}$ & 0.329$_{\pm0.020}$ & \textbf{0.470}$_{\pm0.024}$ \\
 &  & FT & 0.255$_{\pm0.002}$ & 0.244$_{\pm0.005}$ & 0.295$_{\pm0.017}$ & \textbf{0.322}$_{\pm0.021}$ & 0.487$_{\pm0.026}$ \\
 &  & GR & \textbf{0.253}$_{\pm0.001}$ & 0.237$_{\pm0.007}$ & 0.291$_{\pm0.019}$ & 0.324$_{\pm0.021}$ & 0.484$_{\pm0.026}$ \\
 &  & sGR & 0.256$_{\pm0.017}$ & 0.237$_{\pm0.017}$ & 0.294$_{\pm0.017}$ & 0.323$_{\pm0.021}$ & 0.484$_{\pm0.026}$ \\
\cmidrule(l){2-8}
 & $\sqrt{e_S e_T}$ & \cellcolor{gmshade}ZS & \cellcolor{gmshade}0.182$_{\pm0.005}$ & \cellcolor{gmshade}0.166$_{\pm0.004}$ & \cellcolor{gmshade}0.532$_{\pm0.013}$ & \cellcolor{gmshade}0.500$_{\pm0.008}$ & \cellcolor{gmshade}0.460$_{\pm0.005}$ \\
 &  & \cellcolor{gmshade}SC & \cellcolor{gmshade}0.724$_{\pm0.169}$ & \cellcolor{gmshade}0.373$_{\pm0.002}$ & \cellcolor{gmshade}0.710$_{\pm0.024}$ & \cellcolor{gmshade}0.535$_{\pm0.016}$ & \cellcolor{gmshade}0.651$_{\pm0.017}$ \\
 &  & \cellcolor{gmshade}FT & \cellcolor{gmshade}0.292$_{\pm0.013}$ & \cellcolor{gmshade}0.233$_{\pm0.019}$ & \cellcolor{gmshade}0.295$_{\pm0.009}$ & \cellcolor{gmshade}0.381$_{\pm0.013}$ & \cellcolor{gmshade}0.650$_{\pm0.018}$ \\
 &  & \cellcolor{gmshade}GR & \cellcolor{gmshade}\textbf{0.139}$_{\pm0.001}$ & \cellcolor{gmshade}\textbf{0.117}$_{\pm0.002}$ & \cellcolor{gmshade}\textbf{0.205}$_{\pm0.008}$ & \cellcolor{gmshade}\textbf{0.335}$_{\pm0.012}$ & \cellcolor{gmshade}\textbf{0.362}$_{\pm0.010}$ \\
 &  & \cellcolor{gmshade}sGR & \cellcolor{gmshade}0.282$_{\pm0.010}$ & \cellcolor{gmshade}0.120$_{\pm0.005}$ & \cellcolor{gmshade}0.223$_{\pm0.008}$ & \cellcolor{gmshade}0.349$_{\pm0.013}$ & \cellcolor{gmshade}0.368$_{\pm0.010}$ \\
\bottomrule
\end{tabular}

\end{minipage}\par

\FloatBarrier
\clearpage
\section{\GeoReplay{} Budget Sweep}
\label{app:georeplay-budget}
In Table~\ref{tab:georeplay-budget}, we compare \GeoReplay{} and sparse \GeoReplay{} with ordinary fine-tuning and training from scratch across five surrogates and target training budgets $\Beta\in\{1,10,40,80\}$ ($79$ for SHIFT-SUV at the largest budget). We lay out Tables~\ref{tab:georeplay-budget} (pressure), \ref{tab:georeplay-budget-wss} (wall shear stress), \ref{tab:georeplay-budget-drag} (drag), and~\ref{tab:georeplay-budget-lift} (lift) like Table~\ref{tab:main-b10}, grouping rows by dataset, budget, and surrogate and reporting the DrivAerML source error, the target error, and their geometric mean for zero-shot prediction and the four adapted methods. We report SAV in Appendix~\ref{app:results}. We select checkpoints using target validation errors and evaluate them on the fixed held-out test simulations.

{\scriptsize\setlength{\tabcolsep}{2pt}\setlength{\LTcapwidth}{\linewidth}
\begin{longtable}{lrlrrrrrrrrrrrrrrr}
\caption{Pressure errors of \GeoReplay{} across target datasets $T$, target training budgets $\Beta$ and surrogates, laid out as Table~\ref{tab:main-b10}. The blocks give the source error $e_S$ of $\mathrm{M}_T^S$ on the 48 designs of the official DrivAerML validation split, the target error $e_T$ of $\mathrm{M}_T^T$ on the target test simulations, and their geometric mean (shaded). ZS denotes the pretrained model without adaptation, so its errors do not depend on $\Beta$. In each block, bold marks the lowest value in each row.}\label{tab:georeplay-budget}\\
\toprule
& & & \multicolumn{5}{c}{Source error $e_S$} & \multicolumn{5}{c}{Target error $e_T$} & \multicolumn{5}{c}{Geometric mean $\sqrt{e_S\,e_T}$} \\
\cmidrule(lr){4-8}\cmidrule(lr){9-13}\cmidrule(lr){14-18}
$T$ & $\Beta$ & Surrogate & ZS $\downarrow$ & SC $\downarrow$ & FT $\downarrow$ & GR $\downarrow$ & sGR $\downarrow$ & ZS $\downarrow$ & SC $\downarrow$ & FT $\downarrow$ & GR $\downarrow$ & sGR $\downarrow$ & ZS $\downarrow$ & SC $\downarrow$ & FT $\downarrow$ & GR $\downarrow$ & sGR $\downarrow$ \\
\midrule
\endfirsthead
\multicolumn{18}{l}{Table \thetable\ continued}\\
\toprule
& & & \multicolumn{5}{c}{Source error $e_S$} & \multicolumn{5}{c}{Target error $e_T$} & \multicolumn{5}{c}{Geometric mean $\sqrt{e_S\,e_T}$} \\
\cmidrule(lr){4-8}\cmidrule(lr){9-13}\cmidrule(lr){14-18}
$T$ & $\Beta$ & Surrogate & ZS $\downarrow$ & SC $\downarrow$ & FT $\downarrow$ & GR $\downarrow$ & sGR $\downarrow$ & ZS $\downarrow$ & SC $\downarrow$ & FT $\downarrow$ & GR $\downarrow$ & sGR $\downarrow$ & ZS $\downarrow$ & SC $\downarrow$ & FT $\downarrow$ & GR $\downarrow$ & sGR $\downarrow$ \\
\midrule
\endhead
\midrule\multicolumn{18}{r}{Continued on next page}\\\endfoot
\bottomrule\endlastfoot
DrivAerNet++ & 1 & TSol & \textbf{0.042} & 0.789 & 0.313 & 0.231 & 0.306 & 1.020 & 0.504 & 0.347 & \textbf{0.347} & 0.348 & \cellcolor{gmshade}\textbf{0.207} & \cellcolor{gmshade}0.630 & \cellcolor{gmshade}0.329 & \cellcolor{gmshade}0.283 & \cellcolor{gmshade}0.326 \\
 &  & GeoT & \textbf{0.037} & 0.752 & 0.166 & 0.092 & 0.119 & 1.271 & 0.381 & \textbf{0.375} & 0.376 & 0.375 & \cellcolor{gmshade}0.217 & \cellcolor{gmshade}0.535 & \cellcolor{gmshade}0.249 & \cellcolor{gmshade}\textbf{0.186} & \cellcolor{gmshade}0.211 \\
 &  & DoM & \textbf{0.186} & 0.804 & 0.422 & 0.348 & 0.396 & 0.794 & 0.403 & \textbf{0.329} & 0.330 & 0.329 & \cellcolor{gmshade}0.384 & \cellcolor{gmshade}0.569 & \cellcolor{gmshade}0.373 & \cellcolor{gmshade}\textbf{0.339} & \cellcolor{gmshade}0.361 \\
 &  & XMGN & \textbf{0.278} & 0.726 & 0.479 & 0.463 & 0.449 & 1.194 & 0.393 & 0.373 & 0.368 & \textbf{0.367} & \cellcolor{gmshade}0.576 & \cellcolor{gmshade}0.534 & \cellcolor{gmshade}0.423 & \cellcolor{gmshade}0.412 & \cellcolor{gmshade}\textbf{0.406} \\
 &  & GINO & \textbf{0.268} & 0.764 & 0.511 & 0.481 & 0.498 & 0.865 & 0.509 & 0.404 & 0.404 & \textbf{0.404} & \cellcolor{gmshade}0.481 & \cellcolor{gmshade}0.624 & \cellcolor{gmshade}0.454 & \cellcolor{gmshade}\textbf{0.441} & \cellcolor{gmshade}0.449 \\
\cmidrule(l){2-18}
 & 10 & TSol & \textbf{0.042} & 0.736 & 0.350 & 0.096 & 0.290 & 1.020 & 0.215 & 0.197 & \textbf{0.197} & 0.197 & \cellcolor{gmshade}0.207 & \cellcolor{gmshade}0.398 & \cellcolor{gmshade}0.263 & \cellcolor{gmshade}\textbf{0.137} & \cellcolor{gmshade}0.239 \\
 &  & GeoT & \textbf{0.037} & 0.764 & 0.199 & 0.060 & 0.063 & 1.271 & \textbf{0.202} & 0.205 & 0.206 & 0.208 & \cellcolor{gmshade}0.217 & \cellcolor{gmshade}0.393 & \cellcolor{gmshade}0.202 & \cellcolor{gmshade}\textbf{0.111} & \cellcolor{gmshade}0.114 \\
\end{longtable}
}
{\scriptsize\setlength{\tabcolsep}{2pt}\setlength{\LTcapwidth}{\linewidth}
\begin{longtable}{lrlrrrrrrrrrrrrrrr}
\caption{Wall shear stress errors of \GeoReplay{} across target datasets $T$, target training budgets $\Beta$ and surrogates, laid out as Table~\ref{tab:main-b10}. The blocks give the source error $e_S$ of $\mathrm{M}_T^S$ on the 48 designs of the official DrivAerML validation split, the target error $e_T$ of $\mathrm{M}_T^T$ on the target test simulations, and their geometric mean (shaded). ZS denotes the pretrained model without adaptation, so its errors do not depend on $\Beta$. In each block, bold marks the lowest value in each row.}\label{tab:georeplay-budget-wss}\\
\toprule
& & & \multicolumn{5}{c}{Source error $e_S$} & \multicolumn{5}{c}{Target error $e_T$} & \multicolumn{5}{c}{Geometric mean $\sqrt{e_S\,e_T}$} \\
\cmidrule(lr){4-8}\cmidrule(lr){9-13}\cmidrule(lr){14-18}
$T$ & $\Beta$ & Surrogate & ZS $\downarrow$ & SC $\downarrow$ & FT $\downarrow$ & GR $\downarrow$ & sGR $\downarrow$ & ZS $\downarrow$ & SC $\downarrow$ & FT $\downarrow$ & GR $\downarrow$ & sGR $\downarrow$ & ZS $\downarrow$ & SC $\downarrow$ & FT $\downarrow$ & GR $\downarrow$ & sGR $\downarrow$ \\
\midrule
\endfirsthead
\multicolumn{18}{l}{Table \thetable\ continued}\\
\toprule
& & & \multicolumn{5}{c}{Source error $e_S$} & \multicolumn{5}{c}{Target error $e_T$} & \multicolumn{5}{c}{Geometric mean $\sqrt{e_S\,e_T}$} \\
\cmidrule(lr){4-8}\cmidrule(lr){9-13}\cmidrule(lr){14-18}
$T$ & $\Beta$ & Surrogate & ZS $\downarrow$ & SC $\downarrow$ & FT $\downarrow$ & GR $\downarrow$ & sGR $\downarrow$ & ZS $\downarrow$ & SC $\downarrow$ & FT $\downarrow$ & GR $\downarrow$ & sGR $\downarrow$ & ZS $\downarrow$ & SC $\downarrow$ & FT $\downarrow$ & GR $\downarrow$ & sGR $\downarrow$ \\
\midrule
\endhead
\midrule\multicolumn{18}{r}{Continued on next page}\\\endfoot
\bottomrule\endlastfoot
DrivAerNet++ & 1 & TSol & \textbf{0.079} & 0.899 & 0.451 & 0.313 & 0.449 & 1.155 & 0.709 & 0.467 & \textbf{0.467} & 0.468 & \cellcolor{gmshade}\textbf{0.302} & \cellcolor{gmshade}0.798 & \cellcolor{gmshade}0.459 & \cellcolor{gmshade}0.382 & \cellcolor{gmshade}0.458 \\
 &  & GeoT & \textbf{0.070} & 0.948 & 0.283 & 0.152 & 0.194 & 1.019 & \textbf{0.477} & 0.492 & 0.492 & 0.493 & \cellcolor{gmshade}\textbf{0.267} & \cellcolor{gmshade}0.673 & \cellcolor{gmshade}0.373 & \cellcolor{gmshade}0.274 & \cellcolor{gmshade}0.310 \\
 &  & DoM & \textbf{0.280} & 1.248 & 0.550 & 0.472 & 0.525 & 1.091 & 0.512 & \textbf{0.445} & 0.446 & 0.445 & \cellcolor{gmshade}0.553 & \cellcolor{gmshade}0.799 & \cellcolor{gmshade}0.495 & \cellcolor{gmshade}\textbf{0.459} & \cellcolor{gmshade}0.483 \\
 &  & XMGN & \textbf{0.400} & 1.116 & 0.647 & 0.632 & 0.601 & 1.193 & 0.473 & 0.458 & \textbf{0.455} & 0.455 & \cellcolor{gmshade}0.691 & \cellcolor{gmshade}0.727 & \cellcolor{gmshade}0.544 & \cellcolor{gmshade}0.536 & \cellcolor{gmshade}\textbf{0.523} \\
 &  & GINO & \textbf{0.482} & 0.978 & 0.943 & 0.890 & 0.918 & 1.208 & 0.627 & \textbf{0.541} & 0.541 & 0.541 & \cellcolor{gmshade}0.763 & \cellcolor{gmshade}0.783 & \cellcolor{gmshade}0.714 & \cellcolor{gmshade}\textbf{0.694} & \cellcolor{gmshade}0.705 \\
\cmidrule(l){2-18}
 & 10 & TSol & \textbf{0.079} & 0.916 & 0.500 & 0.160 & 0.448 & 1.155 & 0.315 & 0.300 & 0.300 & \textbf{0.299} & \cellcolor{gmshade}0.302 & \cellcolor{gmshade}0.537 & \cellcolor{gmshade}0.387 & \cellcolor{gmshade}\textbf{0.219} & \cellcolor{gmshade}0.366 \\
 &  & GeoT & \textbf{0.070} & 0.923 & 0.318 & 0.114 & 0.119 & 1.019 & 0.315 & 0.307 & \textbf{0.306} & 0.310 & \cellcolor{gmshade}0.267 & \cellcolor{gmshade}0.539 & \cellcolor{gmshade}0.313 & \cellcolor{gmshade}\textbf{0.187} & \cellcolor{gmshade}0.192 \\
\end{longtable}
}
{\fontsize{6}{7.2}\selectfont\setlength{\tabcolsep}{1.5pt}\setlength{\LTcapwidth}{\linewidth}
\begin{longtable}{lrlrrrrrrrrrrrrrrr}
\caption{Drag errors of \GeoReplay{} across target datasets $T$, target training budgets $\Beta$ and surrogates, laid out as Table~\ref{tab:main-b10}. The blocks give the source error $e_S$ of $\mathrm{M}_T^S$ on the 48 designs of the official DrivAerML validation split, the target error $e_T$ of $\mathrm{M}_T^T$ on the target test simulations, and their geometric mean (shaded). ZS denotes the pretrained model without adaptation, so its errors do not depend on $\Beta$. In each block, bold marks the lowest value in each row.}\label{tab:georeplay-budget-drag}\\
\toprule
& & & \multicolumn{5}{c}{Source error $e_S$} & \multicolumn{5}{c}{Target error $e_T$} & \multicolumn{5}{c}{Geometric mean $\sqrt{e_S\,e_T}$} \\
\cmidrule(lr){4-8}\cmidrule(lr){9-13}\cmidrule(lr){14-18}
$T$ & $\Beta$ & Surrogate & ZS $\downarrow$ & SC $\downarrow$ & FT $\downarrow$ & GR $\downarrow$ & sGR $\downarrow$ & ZS $\downarrow$ & SC $\downarrow$ & FT $\downarrow$ & GR $\downarrow$ & sGR $\downarrow$ & ZS $\downarrow$ & SC $\downarrow$ & FT $\downarrow$ & GR $\downarrow$ & sGR $\downarrow$ \\
\midrule
\endfirsthead
\multicolumn{18}{l}{Table \thetable\ continued}\\
\toprule
& & & \multicolumn{5}{c}{Source error $e_S$} & \multicolumn{5}{c}{Target error $e_T$} & \multicolumn{5}{c}{Geometric mean $\sqrt{e_S\,e_T}$} \\
\cmidrule(lr){4-8}\cmidrule(lr){9-13}\cmidrule(lr){14-18}
$T$ & $\Beta$ & Surrogate & ZS $\downarrow$ & SC $\downarrow$ & FT $\downarrow$ & GR $\downarrow$ & sGR $\downarrow$ & ZS $\downarrow$ & SC $\downarrow$ & FT $\downarrow$ & GR $\downarrow$ & sGR $\downarrow$ & ZS $\downarrow$ & SC $\downarrow$ & FT $\downarrow$ & GR $\downarrow$ & sGR $\downarrow$ \\
\midrule
\endhead
\midrule\multicolumn{18}{r}{Continued on next page}\\\endfoot
\bottomrule\endlastfoot
DrivAerNet++ & 1 & TSol & \textbf{0.0107} & 0.5594 & 0.3840 & 0.4019 & 0.3902 & 0.2446 & \textbf{0.1088} & 0.1297 & 0.1303 & 0.1328 & \cellcolor{gmshade}\textbf{0.0512} & \cellcolor{gmshade}0.2467 & \cellcolor{gmshade}0.2232 & \cellcolor{gmshade}0.2288 & \cellcolor{gmshade}0.2276 \\
 &  & GeoT & \textbf{0.0084} & 0.2882 & 0.0323 & 0.0647 & 0.0884 & 0.5001 & \textbf{0.1134} & 0.2248 & 0.2242 & 0.2214 & \cellcolor{gmshade}\textbf{0.0648} & \cellcolor{gmshade}0.1808 & \cellcolor{gmshade}0.0852 & \cellcolor{gmshade}0.1205 & \cellcolor{gmshade}0.1399 \\
 &  & DoM & \textbf{0.0164} & 1.0640 & 0.3589 & 0.3593 & 0.3615 & 0.8529 & 0.1654 & 0.0732 & \textbf{0.0725} & 0.0728 & \cellcolor{gmshade}\textbf{0.1181} & \cellcolor{gmshade}0.4195 & \cellcolor{gmshade}0.1621 & \cellcolor{gmshade}0.1614 & \cellcolor{gmshade}0.1622 \\
 &  & XMGN & \textbf{0.0893} & 0.9603 & 0.2421 & 0.2429 & 0.2615 & 0.3564 & 0.1939 & 0.1055 & 0.1049 & \textbf{0.1044} & \cellcolor{gmshade}0.1784 & \cellcolor{gmshade}0.4316 & \cellcolor{gmshade}0.1598 & \cellcolor{gmshade}\textbf{0.1596} & \cellcolor{gmshade}0.1653 \\
 &  & GINO & \textbf{0.1177} & 0.6088 & 0.2805 & 0.2845 & 0.2822 & 0.2119 & 0.2178 & 0.1596 & \textbf{0.1594} & 0.1595 & \cellcolor{gmshade}\textbf{0.1579} & \cellcolor{gmshade}0.3641 & \cellcolor{gmshade}0.2116 & \cellcolor{gmshade}0.2129 & \cellcolor{gmshade}0.2121 \\
\cmidrule(l){2-18}
 & 10 & TSol & \textbf{0.0107} & 0.4472 & 0.4671 & 0.0435 & 0.4725 & 0.2446 & 0.0483 & 0.0370 & 0.0363 & \textbf{0.0327} & \cellcolor{gmshade}0.0512 & \cellcolor{gmshade}0.1470 & \cellcolor{gmshade}0.1314 & \cellcolor{gmshade}\textbf{0.0398} & \cellcolor{gmshade}0.1243 \\
 &  & GeoT & \textbf{0.0084} & 0.3739 & 0.0783 & 0.0087 & 0.0106 & 0.5001 & 0.0398 & \textbf{0.0211} & 0.0412 & 0.0304 & \cellcolor{gmshade}0.0648 & \cellcolor{gmshade}0.1219 & \cellcolor{gmshade}0.0406 & \cellcolor{gmshade}0.0189 & \cellcolor{gmshade}\textbf{0.0180} \\
\end{longtable}
}
{\scriptsize\setlength{\tabcolsep}{2pt}\setlength{\LTcapwidth}{\linewidth}
\begin{longtable}{lrlrrrrrrrrrrrrrrr}
\caption{Lift errors of \GeoReplay{} across target datasets $T$, target training budgets $\Beta$ and surrogates, laid out as Table~\ref{tab:main-b10}. The blocks give the source error $e_S$ of $\mathrm{M}_T^S$ on the 48 designs of the official DrivAerML validation split, the target error $e_T$ of $\mathrm{M}_T^T$ on the target test simulations, and their geometric mean (shaded). ZS denotes the pretrained model without adaptation, so its errors do not depend on $\Beta$. In each block, bold marks the lowest value in each row.}\label{tab:georeplay-budget-lift}\\
\toprule
& & & \multicolumn{5}{c}{Source error $e_S$} & \multicolumn{5}{c}{Target error $e_T$} & \multicolumn{5}{c}{Geometric mean $\sqrt{e_S\,e_T}$} \\
\cmidrule(lr){4-8}\cmidrule(lr){9-13}\cmidrule(lr){14-18}
$T$ & $\Beta$ & Surrogate & ZS $\downarrow$ & SC $\downarrow$ & FT $\downarrow$ & GR $\downarrow$ & sGR $\downarrow$ & ZS $\downarrow$ & SC $\downarrow$ & FT $\downarrow$ & GR $\downarrow$ & sGR $\downarrow$ & ZS $\downarrow$ & SC $\downarrow$ & FT $\downarrow$ & GR $\downarrow$ & sGR $\downarrow$ \\
\midrule
\endfirsthead
\multicolumn{18}{l}{Table \thetable\ continued}\\
\toprule
& & & \multicolumn{5}{c}{Source error $e_S$} & \multicolumn{5}{c}{Target error $e_T$} & \multicolumn{5}{c}{Geometric mean $\sqrt{e_S\,e_T}$} \\
\cmidrule(lr){4-8}\cmidrule(lr){9-13}\cmidrule(lr){14-18}
$T$ & $\Beta$ & Surrogate & ZS $\downarrow$ & SC $\downarrow$ & FT $\downarrow$ & GR $\downarrow$ & sGR $\downarrow$ & ZS $\downarrow$ & SC $\downarrow$ & FT $\downarrow$ & GR $\downarrow$ & sGR $\downarrow$ & ZS $\downarrow$ & SC $\downarrow$ & FT $\downarrow$ & GR $\downarrow$ & sGR $\downarrow$ \\
\midrule
\endhead
\midrule\multicolumn{18}{r}{Continued on next page}\\\endfoot
\bottomrule\endlastfoot
DrivAerNet++ & 1 & TSol & \textbf{0.256} & 9.188 & 2.352 & 1.253 & 2.357 & 3.285 & 1.701 & 0.500 & \textbf{0.466} & 0.483 & \cellcolor{gmshade}0.917 & \cellcolor{gmshade}3.954 & \cellcolor{gmshade}1.085 & \cellcolor{gmshade}\textbf{0.764} & \cellcolor{gmshade}1.067 \\
 &  & GeoT & \textbf{0.275} & 2.527 & 1.146 & 0.326 & 0.280 & 3.902 & \textbf{2.370} & 5.693 & 5.281 & 5.935 & \cellcolor{gmshade}\textbf{1.035} & \cellcolor{gmshade}2.447 & \cellcolor{gmshade}2.554 & \cellcolor{gmshade}1.313 & \cellcolor{gmshade}1.289 \\
 &  & DoM & \textbf{1.604} & 8.788 & 3.459 & 3.776 & 3.594 & 14.903 & \textbf{1.660} & 4.365 & 4.155 & 4.336 & \cellcolor{gmshade}4.889 & \cellcolor{gmshade}\textbf{3.820} & \cellcolor{gmshade}3.886 & \cellcolor{gmshade}3.961 & \cellcolor{gmshade}3.947 \\
 &  & XMGN & \textbf{2.300} & 17.559 & 11.439 & 11.381 & 11.298 & 57.404 & 3.843 & \textbf{2.886} & 3.028 & 2.945 & \cellcolor{gmshade}11.490 & \cellcolor{gmshade}8.215 & \cellcolor{gmshade}\textbf{5.746} & \cellcolor{gmshade}5.870 & \cellcolor{gmshade}5.768 \\
 &  & GINO & 0.988 & 5.697 & 0.913 & 0.922 & \textbf{0.911} & 4.065 & \textbf{2.269} & 2.680 & 2.684 & 2.690 & \cellcolor{gmshade}2.004 & \cellcolor{gmshade}3.595 & \cellcolor{gmshade}\textbf{1.564} & \cellcolor{gmshade}1.573 & \cellcolor{gmshade}1.565 \\
\cmidrule(l){2-18}
 & 10 & TSol & \textbf{0.256} & 14.422 & 3.983 & 0.536 & 3.386 & 3.285 & 3.272 & 1.278 & \textbf{0.945} & 1.244 & \cellcolor{gmshade}0.917 & \cellcolor{gmshade}6.869 & \cellcolor{gmshade}2.257 & \cellcolor{gmshade}\textbf{0.712} & \cellcolor{gmshade}2.052 \\
 &  & GeoT & \textbf{0.275} & 5.891 & 1.701 & 0.557 & 0.366 & 3.902 & \textbf{2.089} & 3.349 & 2.518 & 2.333 & \cellcolor{gmshade}1.035 & \cellcolor{gmshade}3.508 & \cellcolor{gmshade}2.387 & \cellcolor{gmshade}1.184 & \cellcolor{gmshade}\textbf{0.924} \\
\end{longtable}
}
\noindent\textit{Notes.} Errors are dimensionless relative $L_2$; lower is better. Source entries require all 48 official DrivAerML validation geometries and the same validation-selected checkpoint as the target entries. Tied values at displayed precision share the bold.
\par Sparse \GeoReplay{} attempts replay every fourth global training step, with a hard alignment threshold of 0.02 and replay weight 0.5. Transolver restricts source replay gradients to its final two blocks and uses 25,000 source points; the other surrogates use all layers and their native point sampling. Target gradients update all layers. These variants differ in more than replay frequency.

\FloatBarrier
\section{Complete SAV Results}
\label{app:results}

\paragraph{Surrogate Added Value.}\label{app:sav}
Surrogate Added Value (SAV) compares the adapted model with the same architecture trained from scratch on the same target simulations. For held-out simulation $i$, let $e_{\mathrm{adapt},i}$ and $e_{\mathrm{SC},i}$ denote the errors of the adapted and from-scratch models, and let $\bar e_{\mathrm{adapt}}$ and $\bar e_{\mathrm{SC}}$ denote their means over the $n$ held-out simulations. The adapted model starts from the pretrained surrogate. It is the fine-tuned model (FT) in Table~\ref{tab:mainsav-b10}. We define
\begin{equation}
\mathrm{SAV}=\frac{\bar e_{\mathrm{SC}}-\bar e_{\mathrm{adapt}}}{\bar e_{\mathrm{SC}}}
=1-\frac{\bar e_{\mathrm{adapt}}}{\bar e_{\mathrm{SC}}}.
\label{eq:sav}
\end{equation}
For $\bar e_{\mathrm{SC}}>0$, $\mathrm{SAV}\in(-\infty,1]$, or $(-\infty,100\%]$ when reported as a percentage. A value of 1 represents zero adapted error, zero represents equal errors, and a negative value means that the adapted model has higher error. A value below $-100\%$ means that the adapted error exceeds twice the from-scratch error. SAV is conditional on the source and target datasets, surrogate, predicted quantity, target-data budget, and training protocol. SAV is close to the accuracy and efficiency gains of Poseidon \citep{herde2024poseidon}, which compare a pretrained PDE model with a model trained from scratch at a matched number of training samples. Poseidon's from-scratch baseline is a different architecture (FNO), whereas SAV holds the architecture fixed and therefore isolates the value of pretraining. We measure it after training, so SAV does not by itself predict transfer before adaptation. We form a 95\% interval by resampling the paired held-out simulations and recomputing Equation~\ref{eq:sav}. This interval measures variation across the held-out simulations conditional on the fixed split and trained models. To prevent a few high-error simulations from determining the direction of a claim, we call transfer beneficial or harmful only when the interval excludes zero and the SAV sign agrees with $n^{-1}\sum_i\log(e_{\mathrm{SC},i}/e_{\mathrm{adapt},i})$.

As a reference, we report the best per-simulation mean-field predictor in the Mean column of Table~\ref{tab:mainsav-b10}, which we obtain by predicting each held-out geometry's own spatial-mean field. No mean-field predictor achieves lower error, so we read a model above this value as capturing no spatial structure beyond the field mean.

\paragraph{AB-UPT.} AB-UPT \citep{alkin2025ab} needs CFD volume meshes, so Table~\ref{tab:mainsav-b10} includes it only
for AhmedML. On AhmedML we also trained it with $\Beta=1$, 4, 20, 40 and 150 target simulations, on the
splits of the budget sweep (Table~\ref{tab:abupt-budget}). Fine-tuning beats training from scratch with $\Beta=1$ and
$\Beta=4$ (SAV $+26.0\%$ and $+7.8\%$ in pressure $L_2$ error), loses with $\Beta=10$, 20 and 40 ($-37.4\%$, $-98.4\%$
and $-45.1\%$), and matches it with $\Beta=150$ ($-1.1\%$). We did not run \GeoReplay{} with AB-UPT.

\begin{table*}[!t]
\centering
\caption{At $\Beta=10$ target training simulations, mean test error for the mean-field baseline (Mean), zero-shot source-only prediction (ZS), fine-tuning (FT), and training from scratch (SC), with $\mathrm{SAV}=1-\mathrm{FT}/\mathrm{SC}$, on the target test simulations. Blue and orange shade positive and negative SAV estimates, and gray marks $|\mathrm{SAV}|<1\%$. \textbf{Bold}/\underline{underlined} mark the lowest/second-lowest errors per SC/FT:target results.}
\label{tab:mainsav-b10}
\footnotesize
\setlength{\tabcolsep}{4pt}
\begin{tabular}{llrrrrr@{\hspace{10pt}}rrrrr}
\toprule
& & \multicolumn{5}{c}{Pressure relative $L_2$} & \multicolumn{5}{c}{Wall shear stress relative $L_2$} \\
\cmidrule(lr){3-7}\cmidrule(lr){8-12}
Target dataset & Surrogate & Mean $\downarrow$ & ZS $\downarrow$ & FT $\downarrow$ & SC $\downarrow$ & SAV (\%) $\uparrow$ & Mean $\downarrow$ & ZS $\downarrow$ & FT $\downarrow$ & SC $\downarrow$ & SAV (\%) $\uparrow$ \\
\midrule
 \multirow{5}{*}{DrivAerNet++} & TSol & \multirow{5}{*}{0.771} & 1.020 & \textbf{0.197} & \underline{0.215} & \savbenefit{$+8.3$} & \multirow{5}{*}{0.893} & 1.155 & \textbf{0.300} & \textbf{0.315} & \savbenefit{$+4.7$} \\
  & GeoT &  & 1.271 & \underline{0.205} & \textbf{0.202} & \savharm{$-1.7$} &  & 1.019 & \underline{0.307} & \textbf{0.315} & \savbenefit{$+2.6$} \\
  & DoM &  & 0.794 & 0.214 & 0.242 & \savbenefit{$+11.6$} &  & 1.091 & 0.317 & 0.342 & \savbenefit{$+7.2$} \\
  & XMGN &  & 1.194 & 0.226 & 0.234 & \savbenefit{$+3.4$} &  & 1.193 & 0.338 & \underline{0.335} & \savinconclusive{$-0.8$} \\
  & GINO &  & 0.865 & 0.283 & 0.283 & \savinconclusive{$-0.1$} &  & 1.208 & 0.386 & 0.380 & \savharm{$-1.6$} \\
\midrule
 \multirow{5}{*}{SHIFT-SUV} & TSol & \multirow{5}{*}{0.892} & 0.733 & \underline{0.097} & \underline{0.103} & \savbenefit{$+5.9$} & \multirow{5}{*}{0.958} & 1.422 & \underline{0.179} & 0.172 & \savharm{$-4.2$} \\
  & GeoT &  & 0.845 & \textbf{0.081} & \textbf{0.067} & \savharm{$-20.7$} &  & 1.264 & \textbf{0.133} & \textbf{0.100} & \savharm{$-33.6$} \\
  & DoM &  & 0.687 & 0.119 & 0.134 & \savbenefit{$+11.1$} &  & 1.264 & 0.206 & 0.254 & \savbenefit{$+19.0$} \\
  & XMGN &  & 1.110 & 0.117 & \underline{0.103} & \savharm{$-13.7$} &  & 1.937 & 0.198 & \underline{0.167} & \savharm{$-18.5$} \\
  & GINO &  & 0.582 & 0.239 & 0.221 & \savharm{$-8.3$} &  & 1.731 & 0.388 & 0.365 & \savharm{$-6.3$} \\
\midrule
 \multirow{6}{*}{AhmedML} & TSol & \multirow{6}{*}{0.881} & 1.041 & \underline{0.165} & 0.136 & \savharm{$-21.6$} & \multirow{6}{*}{0.700} & 0.876 & \underline{0.133} & \underline{0.116} & \savharm{$-14.3$} \\
  & GeoT &  & 0.988 & \textbf{0.148} & \textbf{0.124} & \savharm{$-19.1$} &  & 0.800 & \textbf{0.115} & \textbf{0.108} & \savharm{$-6.2$} \\
  & DoM &  & 1.242 & 0.256 & 0.243 & \savharm{$-5.5$} &  & 0.793 & 0.173 & 0.172 & \savinconclusive{$-0.6$} \\
  & XMGN &  & 2.036 & 0.292 & 0.329 & \savbenefit{$+11.3$} &  & 0.910 & 0.208 & 0.234 & \savbenefit{$+11.1$} \\
  & GINO &  & 1.043 & 0.486 & 0.590 & \savbenefit{$+17.6$} &  & 0.775 & 0.369 & 0.443 & \savbenefit{$+16.6$} \\
  & AB-UPT$^\dagger$ &  & 0.971 & 0.173 & \underline{0.126} & \savharm{$-37.4$} &  & 0.827 & 0.152 & 0.124 & \savharm{$-22.7$} \\
\midrule
 \multirow{5}{*}{WindsorML} & TSol & \multirow{5}{*}{0.694} & 0.794 & \underline{0.255} & 0.427 & \savbenefit{$+40.3$} & \multirow{5}{*}{0.547} & 0.877 & \underline{0.193} & 0.296 & \savbenefit{$+34.8$} \\
  & GeoT &  & 0.744 & \textbf{0.244} & \textbf{0.173} & \savharm{$-40.9$} &  & 0.793 & \textbf{0.175} & \textbf{0.131} & \savharm{$-32.9$} \\
  & DoM &  & 1.519 & 0.295 & \underline{0.279} & \savharm{$-5.6$} &  & 1.344 & 0.218 & \underline{0.204} & \savharm{$-6.6$} \\
  & XMGN &  & 0.900 & 0.322 & 0.329 & \savbenefit{$+2.2$} &  & 0.889 & 0.222 & 0.237 & \savbenefit{$+6.2$} \\
  & GINO &  & 0.790 & 0.487 & 0.470 & \savharm{$-3.7$} &  & 0.809 & 0.363 & 0.352 & \savharm{$-3.3$} \\
\bottomrule
\end{tabular}
\par\smallskip
\begin{minipage}{\linewidth}\footnotesize $^\dagger$AB-UPT requires CFD volume meshes, and we report it for AhmedML only.\end{minipage}
\end{table*}

\begin{table}[!h]
\centering
\caption{AB-UPT on AhmedML: mean pressure $L_2$ test error of fine-tuning (FT) and training from scratch (SC) for each
target training budget $\Beta$, and $\mathrm{SAV}=1-\mathrm{FT}/\mathrm{SC}$. There is no $\Beta=80$ run.}
\label{tab:abupt-budget}
\small
\begin{tabular}{rccr}
\toprule
$\Beta$ & FT & SC & SAV \\
\midrule
1 & 0.653 & 0.882 & \savbenefit{$+26.0$} \\
4 & 0.299 & 0.325 & \savbenefit{$+7.8$} \\
10 & 0.173 & 0.126 & \savharm{$-37.4$} \\
20 & 0.108 & 0.054 & \savharm{$-98.4$} \\
40 & 0.084 & 0.058 & \savharm{$-45.1$} \\
150 & 0.067 & 0.066 & \savharm{$-1.1$} \\
\bottomrule
\end{tabular}
\end{table}

\FloatBarrier
\section{Proofs}
\label{app:proofs}

\begin{proof}[Proof of Proposition~\ref{prop:risk-bound}]
For the fixed parameters $\theta$, write $R_T(\theta)=\mathbb E_{G\sim P_T^G}[m_T(G)]$ and $R_{S,w}(\theta)=\mathbb E_{G\sim Q_w^G}[m_S(G)]$. Add and subtract $\mathbb E_{G\sim Q_w^G}[m_T(G)]$ and apply the triangle inequality:
\begin{align}
|R_T-R_{S,w}|
&\leq \left|\mathbb E_{P_T^G}m_T-\mathbb E_{Q_w^G}m_T\right|
+\left|\mathbb E_{Q_w^G}(m_T-m_S)\right|.
\end{align}
For any coupling $\pi\in\Pi(P_T^G,Q_w^G)$, the first term is at most $\mathbb E_{(G,G')\sim\pi}|m_T(G)-m_T(G')|\leq L_G\mathbb E_\pi d_{\mathrm C}(G,G')$. Taking the infimum over couplings gives $L_G\mathcal T_{d_{\mathrm C}}(P_T^G,Q_w^G)$. The conditional-shift assumption bounds the second term by $\delta_{\mathrm{phys}}$. Adding the two bounds proves Equation~\ref{eq:risk-bound}.
\end{proof}

The first assumption requires nearby normalized geometries to have nearby conditional target risks for the fixed surrogate. The conditional-shift assumption does not require source and target CFD labels to agree. Instead, $\delta_{\mathrm{phys}}$ explicitly permits disagreement caused by factors that $d_{\mathrm C}$ omits. A small transport term is therefore insufficient when the source and target simulation processes differ substantially.

\begin{proof}[Proof of Proposition~\ref{prop:descent}]
The $L$-smoothness of $F_T$ gives the descent lemma
\begin{align}
F_T(\theta-\eta d)
&\leq F_T(\theta)-\eta\langle g_T,d\rangle+\frac{L\eta^2}{2}\lVert d\rVert^2.
\end{align}
The alignment condition and $\lambda\geq0$ imply
\begin{equation}
\langle g_T,d\rangle=\lVert g_T\rVert^2+\lambda\langle g_T,g_R\rangle\geq\lVert g_T\rVert^2\geq0.
\end{equation}
When $g_T\neq0$, substituting the step-size condition in Equation~\ref{eq:step-condition} into the descent lemma makes the sum of its last two terms nonpositive. When $g_T=0$, Equation~\ref{eq:aligned-gradient} gives $g_R=0$, so the update leaves $\theta$ unchanged. Thus $F_T(\theta^+)\leq F_T(\theta)$ in both cases.
\end{proof}

The proposition uses the exact target empirical gradient. With stochastic mini-batches, the same calculation applies conditionally to the sampled target loss, while a guarantee for the full empirical risk additionally requires assumptions on gradient noise. We therefore interpret the gate as a local safeguard and make no guarantee of improved generalization.

\FloatBarrier
\section{Implementation Details}
\label{app:implementation}

\paragraph{Source Model Training.} From DrivAerML we use the 436 simulations of the official training split, which pretrained the released models, and the 48 of the official validation split for source evaluation. We use all 500 DrivAerNet++ simulations and the 99 SHIFT-SUV sample simulations. We use 499 of the 500 AhmedML simulations and exclude run 500, whose released force file is unavailable, so its drag cannot be checked against the integrated surface fields. For WindsorML we use the official training, validation, and test splits of the release. It lists 355 designs, but runs 350 to 354 have no published fields, which leaves 280 training, 35 validation, and 35 test simulations, and each training budget is a nested subset of the official training split.

We use the model implementations and released DrivAerML checkpoints from PhysicsNeMo-CFD version 0.0.3a0 with PhysicsNeMo 2.1.1 \citep{tangsali2025benchmarking}. Each matched pair uses the same dataset adapter, surface sampling, architecture, loss, optimizer family, split, epoch budget, and checkpoint-selection rule, and differs only in initialization and in an initialization-appropriate learning rate. The fine-tuned arm begins from the released source parameters; the scratch arm begins from random initialization. We train each arm to convergence with seed 0. Each run stops after 30 consecutive epochs without a validation improvement or at an epoch cap of 300, raised to 500 for arms whose validation error was still falling at 300 epochs. We retain each surrogate's native learning-rate schedule: Transolver and GeoTransolver use step decay every 100 epochs, DoMINO uses its released multistep schedule, and X-MeshGraphNet uses cosine decay. We reached the 300-epoch cap in stages by resuming the saved model, optimizer, scheduler, and early-stopping state; both arms of every pair follow the identical schedule. The validation score is the sum of mean pressure and wall shear stress relative $L_2$ errors. We evaluate the test simulations once, using the checkpoint with the lowest validation score over the whole run.

The fine-tuned arm uses learning rate $5\times10^{-5}$, suited to continuing from pretrained parameters, and the scratch arm uses $1\times10^{-3}$, because a randomly initialized model needs a larger rate and reusing the small fine-tuning rate would understate scratch performance and inflate SAV. Transolver trains in bfloat16 precision on sampled surface points; its released 200,000-point setting exceeds the memory of one 140 GB H200 during a backward pass. It samples 100,000 points per simulation for the first 100 epochs on H200 GPUs and 64,000 points afterwards on 80 GB H100 GPUs; both arms of each pair use the same sampling, and evaluation always covers the complete surface. The other architectures follow their native PhysicsNeMo surface representations and sampling conventions. GINO follows the published Ahmed-body configuration of \citet{li2023geometry}: a $64^3$ latent grid on the normalized domain $[-1,1]^3$, a GNO radius of 0.055, 64 FNO channels, a signed distance field of the geometry as latent input, and 100,000 sampled surface points. We pretrain it on the 436 DrivAerML training simulations for 100 epochs with Adam and a step schedule that halves the learning rate of $2.5\times10^{-4}$ at epoch 50, and it predicts pressure and wall shear stress at every surface cell. AB-UPT uses the released DrivAerML checkpoint and the released sampling configuration of 65,536 geometry points, 16,384 supernodes, and 16,384 surface and volume anchors. Its volume anchors are a uniform subsample of the cell centers of each geometry's own CFD volume mesh; with the released DrivAerML cloud, this subsample reproduces the released zero-shot surface error on a held-out design. Both arms of AB-UPT follow its released Lion recipe, including an exponential moving average of the weights over a 25-epoch window. The complete campaign contains 40 training arms: fine-tuned and scratch runs for five architectures and four datasets. We separately run 20 zero-shot evaluations on the exact test simulations.

Before training or inference, we reconcile each dataset with the source coordinate, scale, and field-unit conventions. WindsorML uses a vertical $y$ and lateral $z$ axis and releases dimensionless pressure and friction coefficients on mesh points; AhmedML releases kinematic fields computed at a unit freestream speed; SHIFT-SUV contains quarter-scale vehicles. We convert point fields to cell fields, redimensionalize WindsorML coefficients and AhmedML kinematic fields at the dynamic pressure of the released PhysicsNeMo-CFD model wrappers ($\rho=1.205$\,kg/m$^3$, $U=30$\,m/s), scale SHIFT-SUV geometry by four and place it on the ground plane, and align normal and wall shear stress sign conventions with the source. We keep the WindsorML axes and body sizes as released, so these frame and scale differences remain part of the shift each surrogate faces. DrivAerNet++ requires no coordinate-scale or field-unit conversion. Force integration after conversion reproduces the available dataset force coefficients; detailed validation is retained in the experiment archive. Every method, including training from scratch, uses the released checkpoint's PhysicsNeMo input preprocessing and output normalization statistics at both training and inference, so methods differ only in initialization and training and never in normalization.

\paragraph{Assets and computing infrastructure.}
DrivAerML \citep{ashton2024drivaerml}, AhmedML \citep{ashton2024ahmedml}, and WindsorML \citep{ashton2024windsorml} are released under CC BY-SA 4.0, and DrivAerNet++ \citep{elrefaie2024drivaernet++} and the SHIFT-SUV sample \citep{shift_suv_2025} under CC BY-NC 4.0. The DrivAerML checkpoints of Transolver, GeoTransolver, DoMINO, and X-MeshGraphNet are released under the NVIDIA Open Model License, PhysicsNeMo and PhysicsNeMo-CFD \citep{tangsali2025benchmarking} under Apache 2.0, the GINO implementation \citep{li2023geometry} under the MIT License, and AB-UPT \citep{alkin2025ab} under the Emmi AI Non-Production License. We use all assets for non-commercial research. We train and evaluate on an internal cluster with NVIDIA H200 and RTX A6000 GPUs and on a cloud cluster with NVIDIA H100 80\,GB GPUs.

\paragraph{Sparse \GeoReplay{}.}
We also evaluate a sparse variant of \GeoReplay{} that samples one source simulation from the same geometry-weighted distribution every fourth target training step and uses 25,000 source surface points. We restrict the source gradient to the final two Transolver blocks, including the output projection, and accept it only when its cosine similarity with the target gradient over those parameters exceeds 0.02. For an accepted source gradient, we match its norm to the target gradient norm over the same parameters and apply a weight of 0.5. Target gradients update all trainable parameters at every step, including steps when we skip or reject source replay.

\FloatBarrier
\section{Additional Qualitative Results}
\label{app:qualitative}
\begin{figure*}[!ht]
\centering
\includegraphics[width=\linewidth]{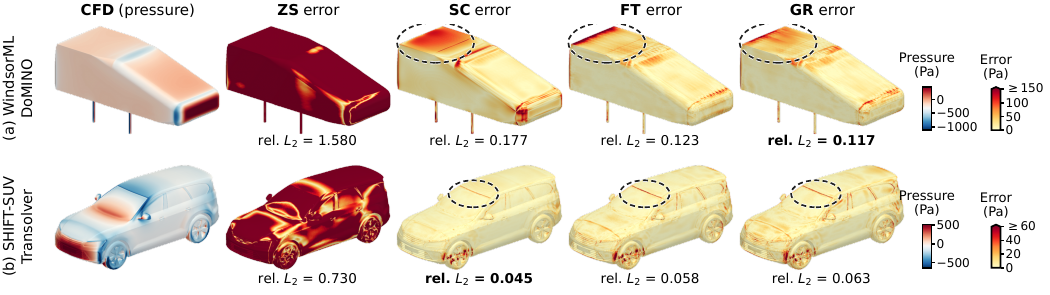}
\caption{Each row shows the test geometry nearest the median per-simulation pressure advantage of fine-tuning over training from scratch, at $\Beta=80$ ($\Beta=79$ for SHIFT-SUV). \textbf{Bold} marks the lowest relative $L_2$ error among SC, FT, and GR, and dashed ellipses mark the same surface region on their maps.}
\label{fig:qualitative}
\end{figure*}

In Figure~\ref{fig:qualitative}, we show the CFD pressure and the absolute pressure error of each method on one held-out geometry per comparison. On WindsorML, fine-tuning DoMINO removes most of the error over the rear half of the roof that remains after training from scratch, and \GeoReplay{} has the lowest error of the three trained models. On SHIFT-SUV, fine-tuning Transolver leaves more error than training from scratch along the leading edge of the roof, and \GeoReplay{} does not close this gap. A large reduction of zero-shot error can therefore coexist with negative transfer value.

\FloatBarrier
\section{Mixed-Dataset Training}
\label{app:pretraining}

The main experiments adapt surrogates pretrained on DrivAerML alone. As a first step toward pretraining across datasets, we train Transolver \citep{pmlr-v235-wu24r} and GeoTransolver \citep{adams2025geotransolver} from random initialization on the union of DrivAerML, DrivAerNet++, SHIFT-SUV, AhmedML, and WindsorML, and compare each mixed model with a reference of the same architecture trained on a single dataset. The mixture draws each dataset as often as its reference does, so both models see the designs of a dataset equally often. Both are evaluated on the test designs of that dataset without further training (Table~\ref{tab:pt-inmix}). These models use splits and a training recipe different from those of the main experiments, so the errors are not directly comparable with the main tables.

Mixed training lowers the pressure error on DrivAerML by 12\% for Transolver and 13\% for GeoTransolver, and by 1\% and 2\% on DrivAerNet++. On SHIFT-SUV, AhmedML, and WindsorML, the mixed models stay within 6\% of their references in either direction, a difference comparable with the spread of up to 4\% between the two Transolver seeds. Training on all five datasets therefore helps on the two DrivAer-derived datasets and matches single-dataset training on the others. This benefit requires the evaluated dataset to be part of the mixture. In an earlier leave-one-dataset-out experiment, models evaluated on a dataset excluded from their training had pressure errors 6 to 20 times those of the reference trained on that dataset, so mixed training alone did not transfer to an unseen vehicle dataset.

\begin{table}[!h]
\centering
\caption{\textbf{In-domain evaluation at equal exposure.} Pressure $L_2$ error on the test designs of each dataset (the official split where a dataset publishes one) for the reference trained on that dataset only (Ref) and the model trained on all five datasets (Mix), with $\Delta=\mathrm{Mix}/\mathrm{Ref}-1$; negative $\Delta$ means that the mixture is more accurate. The mixture draws each dataset as often as its reference. All models train in bfloat16 precision and are evaluated after the last of 501 epochs. Transolver values are the mean of two seeds (spread at most 4\%).}
\label{tab:pt-inmix}
\small
\begin{tabular}{llrrr}
\toprule
Dataset & Model & Ref & Mix & $\Delta$ \\
\midrule
DrivAerML    & Transolver    & 0.055 & 0.048 & $-$12\% \\
             & GeoTransolver & 0.046 & 0.040 & $-$13\% \\
\midrule
DrivAerNet++ & Transolver    & 0.175 & 0.172 & $-$1\% \\
             & GeoTransolver & 0.169 & 0.166 & $-$2\% \\
\midrule
SHIFT-SUV    & Transolver    & 0.050 & 0.051 & +1\% \\
             & GeoTransolver & 0.047 & 0.048 & +2\% \\
\midrule
AhmedML      & Transolver    & 0.035 & 0.035 & +1\% \\
             & GeoTransolver & 0.031 & 0.033 & +6\% \\
\midrule
WindsorML    & Transolver    & 0.083 & 0.082 & $-$1\% \\
             & GeoTransolver & 0.080 & 0.083 & +5\% \\
\bottomrule
\end{tabular}
\end{table}

\FloatBarrier
\end{document}